\documentclass[sigconf]{acmart}
\usepackage{graphicx}
\usepackage{amsmath}
\usepackage{amsthm}
\usepackage{amssymb}
\usepackage{amsfonts}
\usepackage{bm}
\usepackage{caption}
\usepackage{multirow}
\usepackage[english]{babel}
\usepackage{adjustbox}
\usepackage{subfigure}
\usepackage{booktabs}
\usepackage{array}
\usepackage{algorithm}
\usepackage{algorithmic}
\theoremstyle{definition}
\newtheorem{defn}{Definition}
\newtheorem{theorem}{Theorem}
\newtheorem{lem}{Lemma}
\newtheorem{corollary}{Corollary}

\AtBeginDocument{%
  \providecommand\BibTeX{{%
    \normalfont B\kern-0.5em{\scshape i\kern-0.25em b}\kern-0.8em\TeX}}}

\copyrightyear{2020}
\acmYear{2020}
\setcopyright{acmcopyright}\acmConference[KDD '20]{Proceedings of the 26th ACM
SIGKDD Conference on Knowledge Discovery and Data Mining}{August 23--27,
2020}{Virtual Event, CA, USA}
\acmBooktitle{Proceedings of the 26th ACM SIGKDD Conference on Knowledge
Discovery and Data Mining (KDD '20), August 23--27, 2020, Virtual Event, CA, USA}
\acmPrice{15.00}
\acmDOI{10.1145/3394486.3403150}
\acmISBN{978-1-4503-7998-4/20/08}

\begin{document}
\fancyhead{}
\title{Graph Structural-topic Neural Network}


\author{Qingqing Long}
\authornote{These authors contributed equally to the work.}
\affiliation{
  \institution{Key Laboratory of Machine Perception (Ministry of Education), Peking University, China}
}

 \email{qingqinglong@pku.edu.cn}
 
 \author{Yilun Jin}
 \authornotemark[1]
\affiliation{%
  \institution{The Hong Kong University of Science and Technology}
  \city{Hong Kong SAR, China}
  }
  \email{yilun.jin@connect.ust.hk}

\author{Guojie Song}
\authornote{Corresponding Author.}
\affiliation{%
  \institution{Key Laboratory of Machine Perception (Ministry of Education), Peking University, China}
}
\email{gjsong@pku.edu.cn}

\author{Yi Li}
\affiliation{%
  \institution{Peking University, China}
}
\email{liyi2015@pku.edu.cn}

\author{Wei Lin}
\affiliation{%
  \institution{Alibaba Group}
 }
\email{yangkun.lw@alibaba-inc.com}

\begin{abstract}
Graph Convolutional Networks (GCNs) achieved tremendous success by effectively gathering local features for nodes. However, commonly do GCNs focus more on node features but less on graph structures within the neighborhood, especially higher-order structural patterns. However, such local structural patterns are shown to be indicative of node properties in numerous fields. In addition, it is not just single patterns, but the distribution over all these patterns matter, because networks are complex and the neighborhood of each node consists of a mixture of various nodes and structural patterns. Correspondingly, in this paper, we propose \textit{Graph Structural-topic Neural Network}, abbreviated GraphSTONE \footnote{Codes and datasets are available at https://github.com/YimiAChack/GraphSTONE/}, 
a GCN model that utilizes topic models of graphs, such that the structural topics capture indicative graph structures broadly from a probabilistic aspect rather than merely a few structures. Specifically, we build topic models upon graphs using anonymous walks and Graph Anchor LDA, an LDA variant that selects significant structural patterns first, so as to alleviate the complexity and generate structural topics efficiently. In addition, we design multi-view GCNs to unify node features and structural topic features and utilize structural topics to guide the aggregation. We evaluate our model through both quantitative and qualitative experiments, where our model exhibits promising performance, high efficiency, and clear interpretability. 
\end{abstract}


\maketitle

\section{Introduction}
Graphs\footnote{In this paper we interchangeably use terms \textit{network} and \textit{graph}. } are intractable due to their irregularity and sparsity. Fortunately, Graph Convolutional Networks (GCNs) succeed in learning deep representations of graph vertices and attract tremendous attention due to their performance and scalability. 

While GCNs succeed in extracting local features from a node's neighborhood, it should be noted that they primarily focus on node features and are thus less capable of exploiting local structural properties of nodes. Specifically, uniform aggregation depicts one-hop relations, leaving higher-order structural patterns within the neighborhood less attended. Moreover, it is shown that \cite{oono2020graph} deep GCNs can learn little other than degrees and connected components, which further underscores such inability. However, higher-order local structural patterns of nodes, such as network motifs \cite{milo2002network} do provide insightful guidance towards understanding networks. For example, in social networks, network motifs around a node will shed light on social relationships \cite{granovetter1977strength} and dynamic behaviors \cite{zhou2018dynamic}. 

There have been several works that utilize higher-order structural patterns in GCNs, including \cite{lee2019graph}. However, in \cite{lee2019graph} only a few motifs are selected for each node for convolution, which we consider inadequate. In most cases the higher-order neighborhood of a node consists of nodes with a mixture of characteristics, leading to possibly many structural patterns within the neighborhood. Consequently, selecting a few local structural patterns would be insufficient to fully characterize a node's neighborhood. 

\begin{figure*}[htbp]
\centering
        \includegraphics[width=0.92\linewidth]{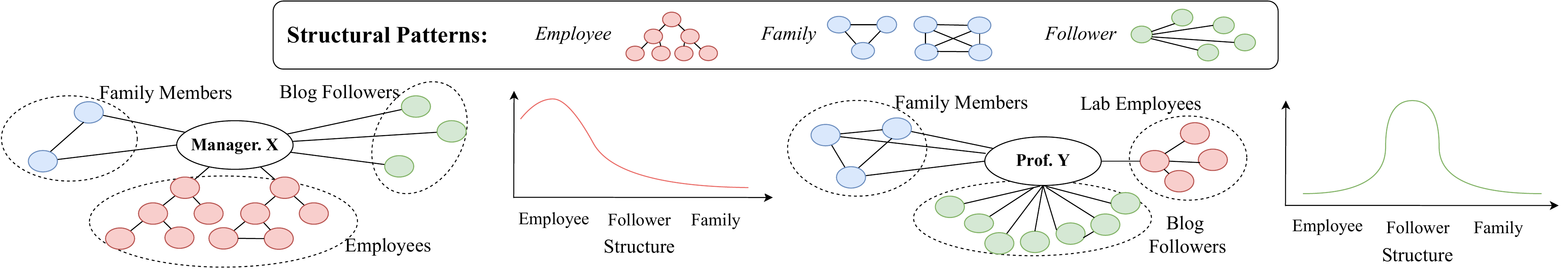}
        \caption{An example of distributional difference of structural patterns in social networks. A manager generally leads a bigger team, while a professor is more influential and is followed by more people. Therefore, while both networks contain the same type of relations and structural patterns, the distributions over them are different.}
        \label{fig:ex_social}
\end{figure*}

We illustrate our claim using Fig. \ref{fig:ex_social} which shows the neighborhoods of a Manager X and a Professor Y, both with three types of relations: family, employees, and followers. Family members know each other well, while employees form hierarchies, and followers may be highly scattered and do not know each other. It can be seen that although both networks contain all three relations, a manager generally leads a larger team, while a professor is more influential and has more followers. As a result, although structural patterns like clusters, trees and stars appear in both neighborhoods, a significant difference in their distributions can be observed. Consequently, it is \textit{the distribution of structural patterns}, rather than individuals, that is required to precisely depict a node's neighborhood. 

Topic modeling is a technique in natural language processing (NLP) where neither documents nor topics are defined by individuals, but distributions of topics and words. Such probabilistic nature immediately corresponds with the \textit{distribution of structural patterns} required to describe complex higher-order neighborhoods of networks. Consequently, we similarly model nodes with \textit{structural topics} to capture such differences in distributions of local structural patterns. For example in Fig. \ref{fig:ex_social}, three structural topics characterized by clusters, trees and stars can be interpreted as family, employees and followers respectively, with Manager X and Prof. Y showing different distributions over the structural topics. 

We highlight two advantages of topic models for graph structural patterns. On one hand, probabilistic modeling captures the distributional differences of local structural patterns for nodes more accurately, which better complements node features captured by GCNs. On the other hand, the structural topics are lower-dimensional representations compared with previous works that directly deal with higher-order structures \cite{DBLP:conf/aaai/JinSS20}, thus possessing less variance and leading to better efficiency.

However, several major obstacles stand in our path towards leveraging topic modeling of structural patterns to enhance GCNs: 
\begin{enumerate}
    \item \textbf{Discovering Structural Patterns} is itself complex. Specifically, previous works \cite{liu2017predicting} generally focus on pre-defined structures, which may not be flexible enough to generalize well on networks with varying nature. Also, many structural metrics require pattern matching, whose time consumption would barely be acceptable for GCNs. 
    \item \textbf{Topic Modeling for Graphs} also requires elaborate effort, as graphs are relational while documents are independent samples. Consequently, adequate adaptations should be made such that the structural topics are technically sound. 
    \item \textbf{Leveraging Structural Features in GCNs} requires unifying node features with structural features of nodes. As they depict different aspects of a node, it would take elaborate designs of graph convolutions such that each set of features would act as a complement to the other. 
\end{enumerate}

In response to these challenges, in this paper we propose \textbf{Graph \underline{S}tructural \underline{To}pic Neural \underline{Ne}twork}, abbreviated \textbf{GraphSTONE}, a GCN framework featuring topic modeling of graph structures. Specifically, we model structural topics via anonymous walks \cite{micali2016reconstructing} and Graph Anchor LDA. On one hand, anonymous walks are a flexible and efficient metric of depicting structural patterns, which only involve sampling instead of matching. On the other hand, we propose Graph Anchor LDA, a novel topic modeling algorithm that pre-selects ``anchors'', i.e. representative structural patterns, which will be emphasized during the topic modeling. By doing so, we are relieved of the overwhelming volume of structural patterns and can thus focus on relatively few key structures. As a result, concise structural topics can be generated with better efficiency. 

We also design multi-view graph convolutions that are able to aggregate node features and structural topic features simultaneously, and utilize the extracted structural topics to guide the aggregation. Extensive experiments are carried out on multiple datasets, where our model outperforms competitive baselines. In addition, we carry out visualization on a synthetic dataset, which provides intuitive understandings of both Graph Anchor LDA and GraphSTONE. 

\begin{figure*}[htbp]
\centering
\includegraphics[width=0.95\linewidth]{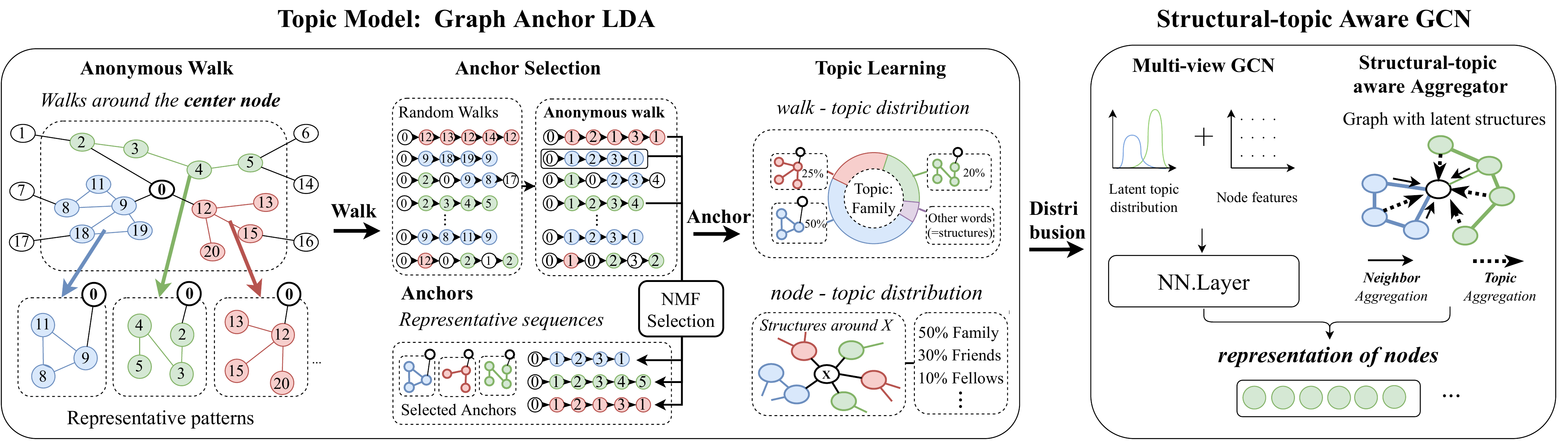}
\caption{An overview of GraphSTONE. GraphSTONE consists of two major components: (a) Graph Anchor LDA, (b) Structural topic aware multi-view GCN.}
\label{fig:framework}
\end{figure*}

To summarize, we make the following contributions.
\begin{enumerate}
    \item We propose structural topic modeling on graphs that capture distributional differences over local structural patterns on graphs, which to the best of our knowledge, is the first attempt to utilize topic modeling on graphs. 
    \item We enable topic modeling on graphs through anonymous walks and a novel Graph Anchor LDA algorithm, which are both flexible and efficient. 
    \item We propose a multi-view GCN unifying both node features with structural topic features, which we show are complementary to each other. 
    \item We carry out extensive experiments on multiple datasets, where GraphSTONE shows competence in both performance and efficiency. 
    
\end{enumerate}

\section{Related Work}
\subsection{Graph Neural Networks (GNNs)} Recent years have witnessed numerous works focusing on deep architectures over graphs \cite{kipf2016semi,hamilton2017inductive}, among which the GCNs received the most attention. GCNs are generally based on neighborhood aggregation, where the computation of a node is carried out by sampling and aggregating features of neighboring nodes. 

Although neighborhood aggregation makes GCNs as powerful as the Weisfeiler-Lehman (WL) isomorphism test \cite{xu2018powerful,morris2019weisfeiler,li2019gcn}, common neighborhood aggregations refer to node features only, leaving them less capable in capturing complex neighborhood structures. Such weakness is also shown in theory. For example, \cite{Loukas2020What} states that GCNs should be sufficiently wide and deep to be able to detect a given subgraph, while \cite{oono2020graph} demonstrates that deep GCNs can learn little other than degrees and connected components. 

To complement, many works have focused on GCNs with emphasis on higher-order local structural patterns. For example, \cite{lee2019graph} selects indicative motifs within a neighborhood before applying attention, which we claim to be insufficient. On the contrary, our work focuses on distributions over structures rather than individual structures. \cite{DBLP:conf/aaai/JinSS20} captures local structural patterns via anonymous walks, which can capture complex structures yet suffers from poor efficiency. By comparison, our solution using topic models would be more efficient in that we pre-select anchors for topic modeling. 



\subsection{Modeling Graph Structures}
There are many previous works on depicting graph structure properties using metrics such as graphlets and shortest paths \cite{shervashidze2009efficient,borgwardt2005shortest}. However, they commonly require pattern matching, which is hardly affordable in large, real-world networks. In addition, these models are constrained to extract pre-designed structural patterns, which are not flexible enough to depict real-world networks with different properties. A parallel line of works, such as \cite{koutra2014vog} aim to decompose a graph into indicative structures. However, they focus on graph-level summarization but fail to generate node-level structural depictions. 

Several works in network embedding also exploit network structures to generate node representations, such as \cite{ribeiro2017struc2vec,donnat2018learning,long2019hierarchical}. However, their focuses are generally singular in that they do not refer to node features, while our model is able to combine both graph structures and node features through GCNs. 

\subsection{Topic Modeling} 
Topic modeling in NLP is a widely used technique aiming to cluster texts. Such models assign a distribution of \textit{topics} to each document, and a distribution of words to each topic to provide low-dimensional, probabilistic descriptions of documents and words. 

Latent Dirichlet Allocation (LDA) \cite{blei2003latent}, a three-level generative model, embodies the most typical topic models. However, although prevalent in NLP \cite{liu2015topical,kawamae2019topic}, LDA has hardly, if ever, been utilized in non-i.i.d. data like networks. In this work, we design a topic model on networks, where structural topics are introduced to capture distributional differences over structural patterns in networks.

\section{Model: GraphSTONE}
In this section we introduce our model \textit{Graph Structural-topic Neural Network}, i.e. \textit{GraphSTONE}. We first present the topic modeling on graphs, before presenting the multi-view graph convolution. 

Fig. \ref{fig:framework} gives an overview of our model GraphSTONE. Anonymous random walks are sampled for each node to depict local structures of a node. Graph Anchor LDA is then carried out on anonymous walks for each node, where we first select ``anchors'', i.e. indicative anonymous walks through non-negative matrix factorization. After obtaining the walk-topic and node-topic distributions, we combine these structural properties with original node features through a multi-view GCN which outputs representations for each node. 

\subsection{Topic Modeling for Graphs}
\subsubsection{Anonymous Walks}
We briefly introduce anonymous walks here and refer readers to \cite{micali2016reconstructing,ivanov2018anonymous} for further details. 

An anonymous walk is similar to a random walk, but with the exact identities of nodes removed. A node in an anonymous walk is represented by the first position where it appears. Fig. \ref{fig:framework} (a) provides intuitive explanations of anonymous walks which we ﬁnd appealing. For example, $w_i=(0,9,8,11,9)$ is a random walk starting from node 0, and its anonymous walk is defined as $w_i=(0,1,2,3,1)$. It is highly likely that it is generated through a triadic closure. 

We present the following theorem to demonstrate the property of anonymous walks in depicting graph structures. 
\begin{theorem} \cite{micali2016reconstructing} Let $B(v,r)$ be the subgraph induced by all nodes $u$ such that $dist(v,u)\leq r$ and $P_L$ be the distribution of anonymous walks of length $L$ starting from $v$, one can reconstruct $B(v,r)$ using $(P_1,...,P_L)$, where $L = 2(m+1)$, $m$ is the number of edges in $B(v,r)$.
\label{thm:anonym}
\end{theorem}

Theorem \ref{thm:anonym} underscores the ability of anonymous walks in describing local structures of nodes in a general manner. Therefore, we take each anonymous walk as a basic pattern for describing graph structures\footnote{Although we do not explicitly reconstruct $B(v, r)$, such theorem demonstrates the ability of anonymous walks to represent structural properties.}. 

\subsubsection{Problem Formulation} We formulate topic modeling on graphs in our paper as follows. 
\begin{defn}[Topic Modeling on Graphs]
Given a graph $G = ( V, E )$, a set of possible anonymous walks of length $l$ as $\mathcal{W}_l$, and the number of desired structural topics $K$,  a topic model on graphs aims to learn the following parameters. 
\begin{itemize}
    \item A node-topic matrix $R\in \mathbb{R}^{|V|\times K}$, where a row $R_i$ corresponds to a distribution with $R_{ik}$ denoting the probability of node $v_i$ belonging to the $k$-th structural topic. 
    \item A walk-topic matrix $U \in \mathbb{R}^{K\times |\mathcal{W}_l|}$ where a row $U_k$ is a distribution over $\mathcal{W}_l$ and $U_{kw}$ denotes the probability of $w\in \mathcal{W}_l$ belonging to the $k$-th structural topic. 
\end{itemize}
In addition, we define the set of anonymous walks starting from $v_i$ as $D_i$, with $D_i = N$ as the number of walks to sample.  
\end{defn}
The formulation is an analogy to topic modeling in NLP, where anonymous walks correspond to \textit{words}, and the sets of walks starting from each node correspond to \textit{documents}. By making the analogy, nodes are given probabilistic depictions over their local structural patterns, and structural topics would thus consist of structural pattern distributions that are indicative towards node properties (social relations in Fig. \ref{fig:ex_social}, for example). 

According to LDA constraints in NLP \cite{arora2012learning}, we introduce Lemma \ref{lem:guarantee} to show that the topic model in networks can indeed be learned.
\begin{lem} 
\label{lem:guarantee}
There is a polynomial-time algorithm that fits a topic model on a graph with error $\epsilon$, if $N$ and the length of walks $l$ satisfy
$$\frac{N}{l} \geq O\Bigg(\frac{b^4K^6}{\epsilon^2 p^6\gamma^2|V|}\Bigg)$$
where $K$ is the number of topics, $|V|$ is the number of nodes. $b, \gamma$ and $p$ are parameters related to topic imbalance defined in \cite{blei2003latent}, which we assume to be fixed. 
\end{lem}

We first introduce the general idea of the lemma. In topic models in NLP, it is assumed that the length of each document $|D_i|$ as well as the vocabulary $\mathcal{W}$ is fixed, while the corpus $|\mathcal{D}|$ is variable-sized. Marked differences exist in graphs, where the number of nodes $|\mathcal{D}| = |V|$ is fixed, while anonymous walk sets and samples are variable-sized. Hence we focus on $N, l$ instead of $|\mathcal{D}|$.  

\begin{proof}
\cite{arora2012learning} gives a lower bound on the number of documents such that the topic model can be fit, namely
$$|\mathcal{D}| = |V| \ge \max\left\{O\left(\frac{\log n \cdot b^4K^6}{\epsilon^2 p^6\gamma^2 N}\right),O\left(\frac{\log K\cdot b^2K^4}{\gamma^2}\right)\right\},$$
where $n = |\mathcal{W}_l|$ is the vocabulary size. As the latter term is constant, we focus on the first term. 

We then get the bound of $N$ and $|\mathcal{W}_l|$, namely
$$\frac{N}{\log |\mathcal{W}_l|}\geq O\left(\frac{b^4K^6}{\epsilon^2 p^6\gamma^2|V|}\right).$$

The number of anonymous walks increases exponentially with length of walks $l$ \cite{ivanov2018anonymous}, i.e., $$\log |\mathcal{W}_l|=\Theta(l). $$

Consequently we reach Lemma \ref{lem:guarantee}.
\end{proof}

\subsubsection{Graph Anchor LDA}
A large number of different walk sequences will be generated on complex networks, among which many may not be indicative, as illustrated in \cite{milo2002network}. If sequences are regarded separately, we would be encountered with a huge ``vocabulary'', which would compromise our model, since the model may overfit on meaningless sequences and ignore more important ones.

Unfortunately, while in NLP the concept of \textit{stopwords} is utilized to remove meaningless words, no such results exist in networks to remove such walk sequences. Consequently, we propose to select highly indicative structures first, which we call ``anchors'', before moving on to further topic modeling. 

Specifically, we define the walk-walk co-occurrence matrix $M\in \mathbb{R}^{|\mathcal{W}_l| \times |\mathcal{W}_l|}$, with $M_{i, j} =\sum_{v_k\in V} \mathbb{I}(w_i \in D_k, w_j\in D_k)$, and adopt non-negative matrix factorization (NMF) \cite{1999learning} to extract anchors
\begin{equation}
\begin{aligned}
&H, Z = \arg\min\|M - HZ\|_F^2\\
&s.t.\quad H, Z^T\in \mathbb{R}^{|\mathcal{W}_l| \times \alpha}, H, Z \ge 0.
\end{aligned}
\label{eq:anchor}
\end{equation}
We iteratively update $H, Z$ until convergence, before finding the anchors by $\mathbf{A}_k=\arg\max(Z_k), k = 1, ...\alpha$, where $\mathbf{A}$ is the set of indices for anchors, and $Z_k$ is the $k$-th row of $Z$. Intuitively, by choosing the walks with largest weights, we are choosing the walks most capable of interpreting the occurrence of other walks, i.e. indicative walks. We later show theoretically that the selected walks are not only indicative of walk co-occurrences but also the underlying topics. 

Based on the anchors we picked, we move forward to learn the walk-topic distribution $U$. \cite{arora2013practical} presents a fast optimization for LDA with anchors as primary indicators and non-anchors providing auxiliary information. We get $U\in\mathbb{R}^{K\times|\mathcal{W}_l|}$ through optimizing
\begin{equation}
\label{eq:lda_2}
\arg\min_U D_{KL}\left(Q_i \| \sum_{k\in \mathbf{A}} U_{ik}\mathrm{diag}^{-1}(Q\vec{1})Q_{\mathbf{A}_k}\right),
\end{equation}
where $Q$ is the re-arranged walk co-occurrence matrix with anchors $\mathbf{A}$ lying in the first $\alpha$ rows and columns, and $Q_{\mathbf{A}_k}$ is the row of $Q$ for the $k$-th anchor. 

In addition, we define node-walk matrix as $Y\in \mathbb{R}^{|V|\times |\mathcal{W}_l|}$ with $Y_{iw}$ denoting the occurrences of $w$ in $D_i$. We then get the node-topic distribution $R$ through $R = YU^\dagger$, where $U^\dagger$ denotes pseudo-inverse.

\subsubsection{Theoretical Analysis}
We here provide a brief theoretical analysis of our Graph Anchor LDA in its ability to recover ``anchors'' of not only walk co-occurrences but also topics. We first formalize the notion of ``anchors'' via the definition of separable matrices. 

\begin{defn}[$p$-separable matrices]
\cite{arora2012learning} An $n × r$ non-negative matrix $C$ is $p$-separable if for each $i$ there is
some row $\pi(i)$ of $C$ that has a single nonzero entry $C_{\pi(i), i}$ with $C_{\pi(i), i}\geq p$.
\end{defn}

Specifically, if a walk-topic matrix $U$ is separable, we call the walks with non-zero weights ``anchors''. We then present a corollary derived from \cite{arora2012learning} indicating that the non-negative matrix factorization is indeed capable of finding such anchors. 

\begin{corollary}
\label{col:anchor}
Suppose the real walk-node matrix (i.e. the real walk distribution of each node) is generated via $Y = U\Lambda$, where $U\Lambda$ is the real walk-topic matrix and $\Lambda$ is a matrix of coefficients, both non-negative. We define $\Sigma = \mathbb{E}[YY^T] = \mathbb{E}[U\Lambda\Lambda^TU^T]$ and $\hat{\Sigma}$ as an observation of $\Sigma$. For every $\varepsilon>0$, there is a polynomial time algorithm that factorizes $\hat{\Sigma} \approx \hat{U}\hat{\Phi}$ such that $\|\hat{U}\hat{\Phi} - \Sigma\|_1\le\varepsilon$. Moreover, if $U$ is $p$-separable, then the rows of $\hat{U}$ almost reconstruct the anchors up to an error of $O(\varepsilon)$.
\end{corollary}

Specifically, the walk co-occurrence matrix $M$ serves as an estimate $\hat{\Sigma}$ of the walk covariance $\Sigma$ in Corollary \ref{col:anchor}. Corollary \ref{col:anchor} emphasizes that our Graph Anchor LDA algorithm is indeed able to select representative structural patterns within each structural topic. In reality, the topics are generally non-separable, where we will empirically show the effectiveness of anchor selection. 

\subsubsection{Discussion}
\begin{figure}[htbp]
\centering

\includegraphics[width=0.5\linewidth]{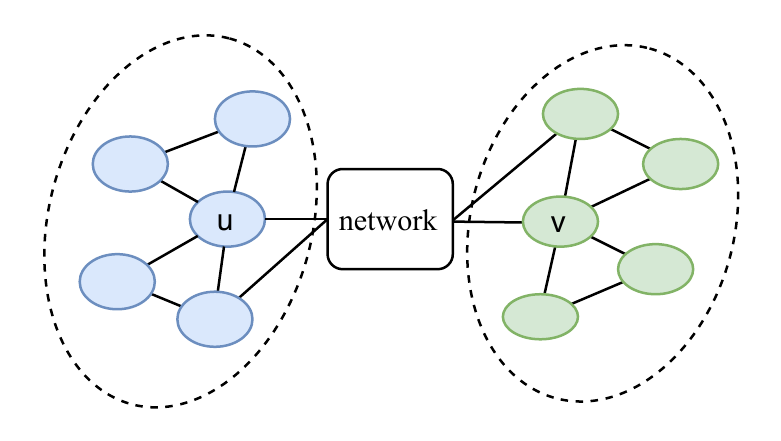}
\caption{An example of two nodes ($u$ and $v$) that are structurally similar, but belongs to distinct communities. }
\label{fig:example_diff_com}
\end{figure}

We discuss the difference between Graph Anchor LDA and community detection, both of which can be used for dimensionality reduction on graphs. While community detection focuses on dense connections, Graph Anchor LDA focuses on the distribution of local structures, which will assign similar topics to structurally similar, but not necessarily connected nodes (e.g. $u$ and $v$ in Fig. \ref{fig:example_diff_com}). We will show such a difference by comparing Graph Anchor LDA with MNMF \cite{jin2019incorporating} in the experiments. 

\begin{algorithm}[tb]
\caption{Algorithm of GraphSTONE}
\label{alg:graphstone}
\begin{algorithmic}[1] 
\REQUIRE Graph $G=(V,E,X)$, number of latent topics $K$
\ENSURE walk-topic distribution matrix $U$, node-topic distribution $R$, node embeddings $\Phi$ with latent topic information  
\STATE $M\leftarrow$ Walk co-occurences($G$)
\STATE Form $\bar{M}=\{\bar{M}_1,\bar{M}_2,...,\bar{M}_V\}$, the normalized rows of $M$.
\STATE $\mathbf{A}\leftarrow$ Graph Anchor LDA ($\bar{M}, K$) (Eq. \ref{eq:anchor}, find anchors)
\STATE $U,R \leftarrow$ RecoverLatentTopic ($M,\mathbf{A}$) (Eq. \ref{eq:lda_2})
\STATE $\Phi\leftarrow$ Structural-topic aware GCN ($G,U,R)$ (Eq. \ref{eq:aggregate}, \ref{eq:multi-view}, and \ref{eq:obj})
\RETURN $U, R,\Phi$
\end{algorithmic}
\end{algorithm}

\subsection{Structural-topic Aware GCN}
We then introduce our design of graph convolutions which fuses structural topic features along with node features.


\subsubsection{Structural-topic Aware Aggregator}

Prevalent GCNs gather all neighbors with equal contributions, while GAT, although assigning different weights on neighbors, takes only node features but not structural properties into account. Consequently, we propose to utilize the learned structural topics to guide the aggregation of neighbors. We utilize the scheme of importance sampling for nodes based on the similarities of their structural topics, such that the neighborhood can be aggregated in a way that illustrates $\textit{homophily}$, i.e. similar nodes influence each other more, 
\begin{equation}
    h_i^{(k)
}=\mathrm{AGGREGATE}\left(\left\{\frac{R_i^TR_j}{\sum_j  R_i^TR_j}h_j^{(k-1)},v_j\in N(v_i)\right\}\right),
\label{eq:aggregate}
\end{equation}
where $h_i^{(k)}$ denotes the output vector for node $v_i$ from the $k$-th layer. We take identical $\mathrm{AGGREGATE}$ as GraphSAGE while being flexible to more sophisticated methods.

\subsubsection{Multi-view GCN}
As the two types of features come from different domains, utilizing one to complement the other would be ideal. Inspired by the idea of boosting, we introduce two parallel GCNs, one focusing on structural topics and the other on node features correspondingly. 
Specifically, let $h_{i,n}^{(k)}$ and $h_{i,s}^{(k)}, k = 1, ..., L$ denote the outputs of the two GCNs for node $v_i$ at layer $k$, we apply a nonlinear neural network layer on the two output vectors.
\begin{equation}
h_i^{(L)}=(\mathbf{W}\cdot \mathrm{tanh}([h_{i,n}^{(L)}\otimes h_{i,s}^{(L)}])+\mathbf{b})
\label{eq:multi-view}
\end{equation} 
where $h_i^{(L)}$ is the final output of the multi-view GCN for node $v_i$, and $\otimes$ can be arbitrary operations, where we take as concatenation. 

For the initialization of $h_i^{(0)}$, we take $h_{i, n}^{(0)}=X_i$, the node feature of $v_i$, and $h_{i, s}^{(0)}$ be the concatenation of the following: a) the vector indicating the distribution of anchors $\mathbf{A}$ in the neighborhood of $v_i$, and b) the node-topic distribution $R_i$.  

Finally, we adopt the unsupervised objective of GraphSAGE \cite{hamilton2017inductive} for learning the output vectors. 
\begin{equation}\mathcal{L}=-\log \left[\sigma\left(h_i^{(L)T}h_j^{(L)}\right)\right]-q\cdot \mathbb{E}_{v_n\sim P_n(v)}\log\left[\sigma\left(h_i^{(L)T}h_{n}^{(L)}\right)\right]
\label{eq:obj}
\end{equation}
where $\sigma(x)$ is the sigmoid function, $v_j$ co-occurs with $v_i$ in random walks, and $P_{n}(v)$ is the noise distribution for negative sampling. 

We show the pseudo-code of GraphSTONE in Algorithm \ref{alg:graphstone}.

\begin{figure*}
        \centering
        \subfigure[Illustration of $G(n)$]{ \includegraphics[width=0.2\linewidth]{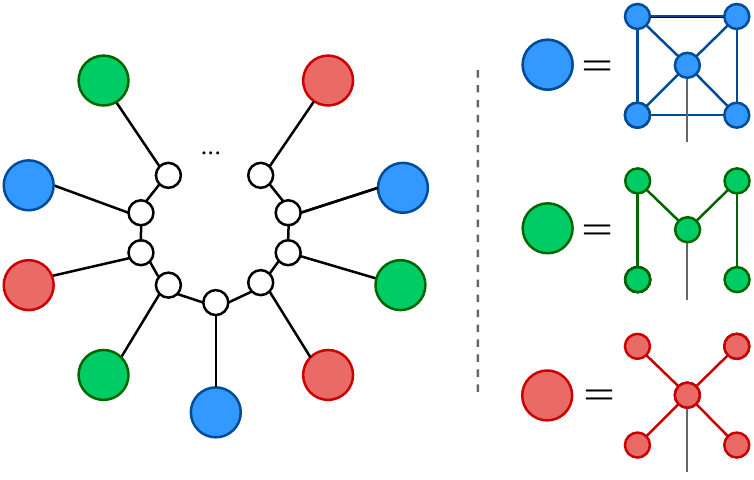}\label{fig:syn_original}}
         \subfigure[Graph Anchor LDA]{ \includegraphics[width=0.17\linewidth]{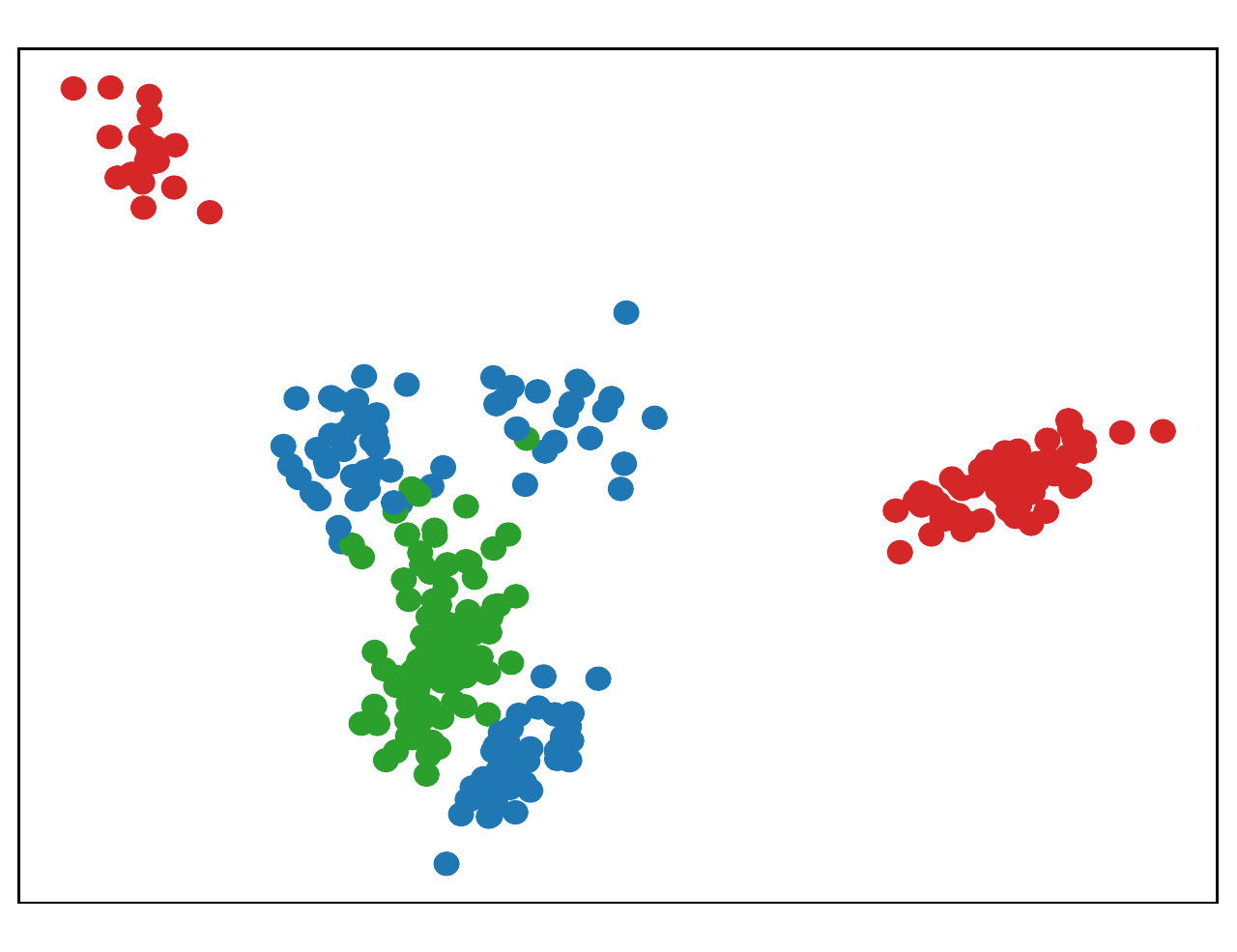}\label{fig:lda_anchor}}
         \subfigure[GraphSTONE]{ \includegraphics[width=0.17\linewidth]{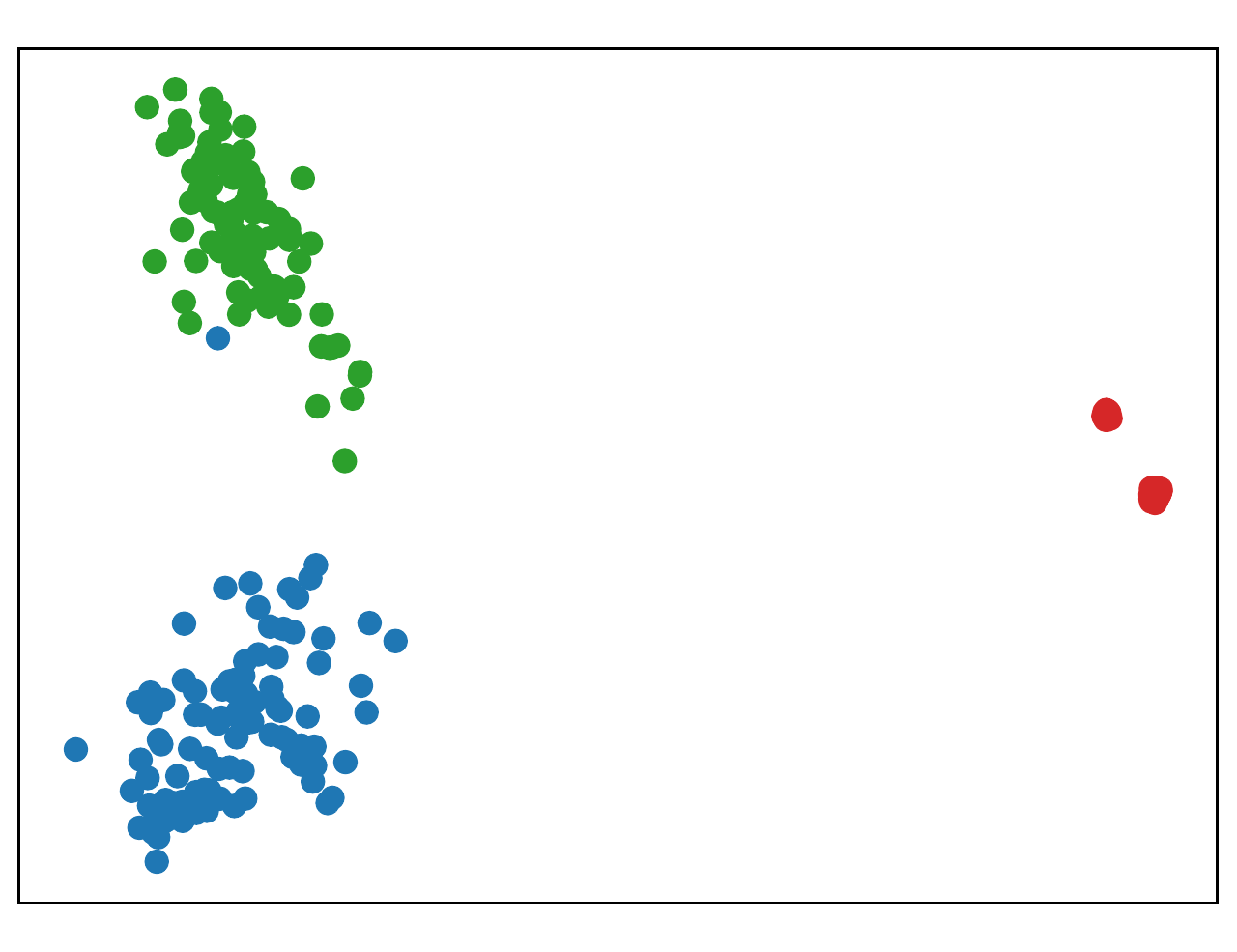}\label{fig:lda_GraphSTONE}}
         \subfigure[GraLSP]{ \includegraphics[width=0.17\linewidth]{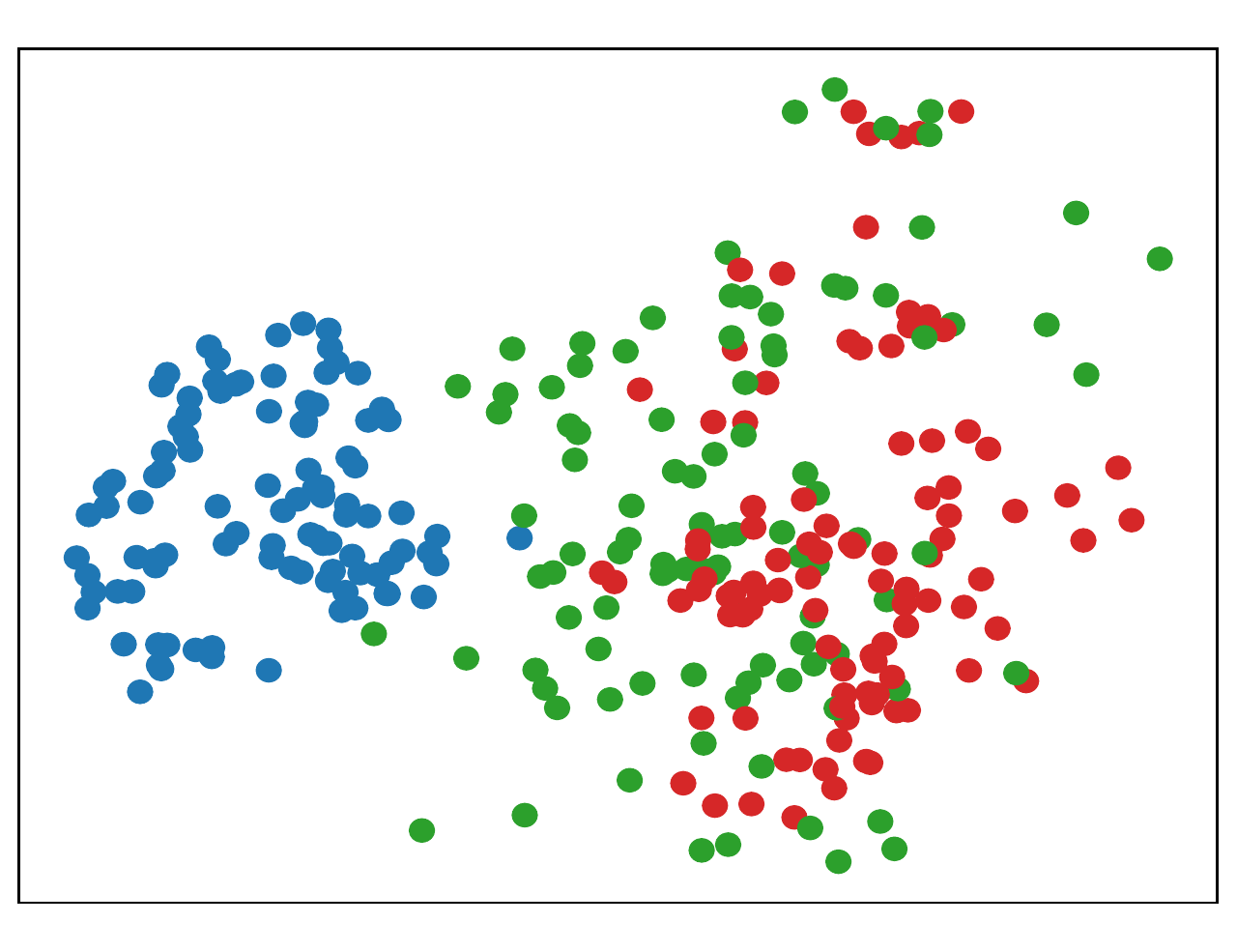}\label{fig:lda_GraLSP}}
         \subfigure[MNMF
        ]{ \includegraphics[width=0.17\linewidth]{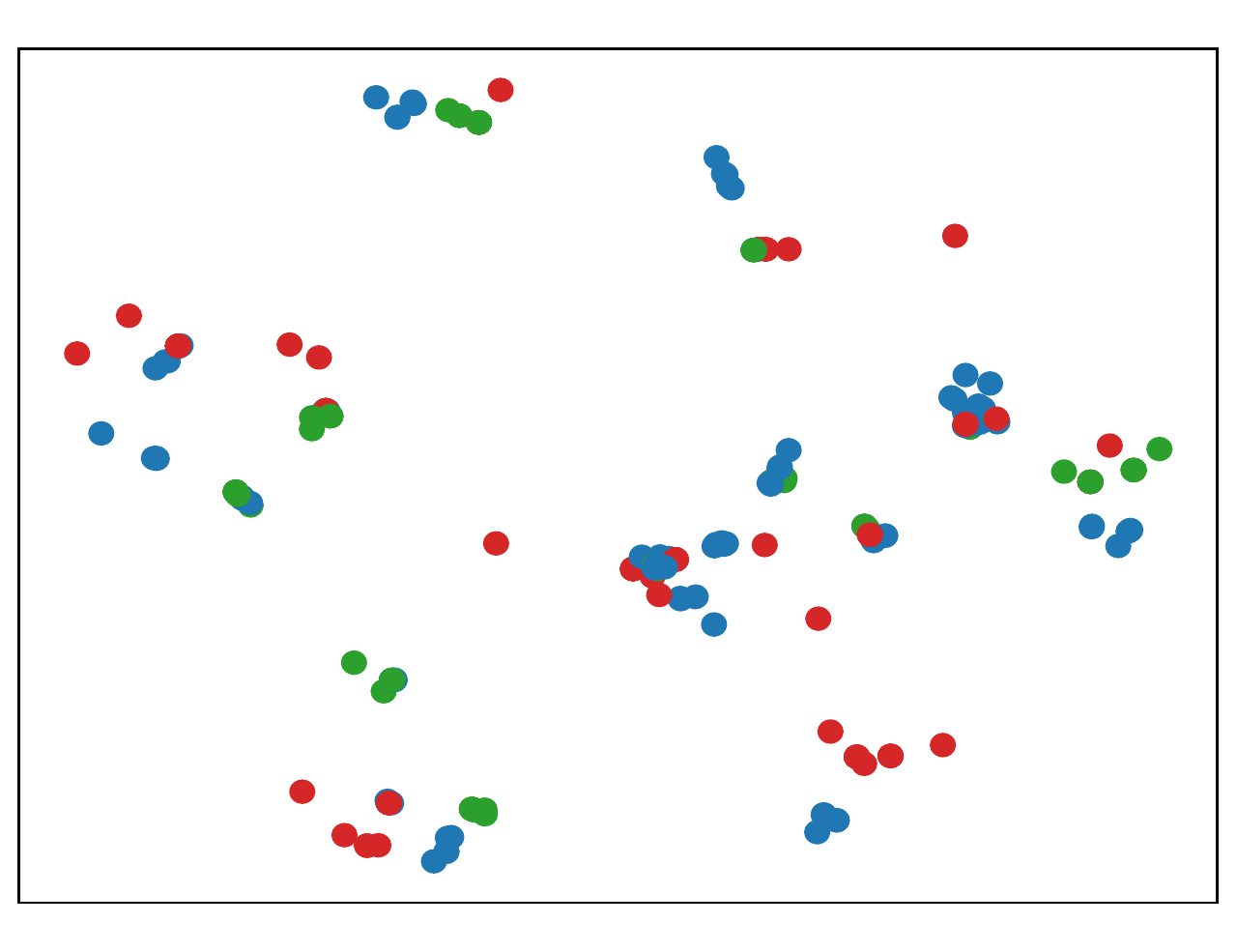}\label{fig:syn_MNMF}
        }
    \caption{Visualization of structural topics, and results by various models on $G(n)$. Graph Anchor LDA and GraphSTONE are able to more clearly mark the differences between local structural patterns than GraLSP and MNMF. }
    \label{fig:show_lda}
\end{figure*}

\begin{figure*}[htbp]
\centering
        \subfigure[Walk-topic distribution by Graph Anchor LDA]{ \includegraphics[width=0.46\linewidth]{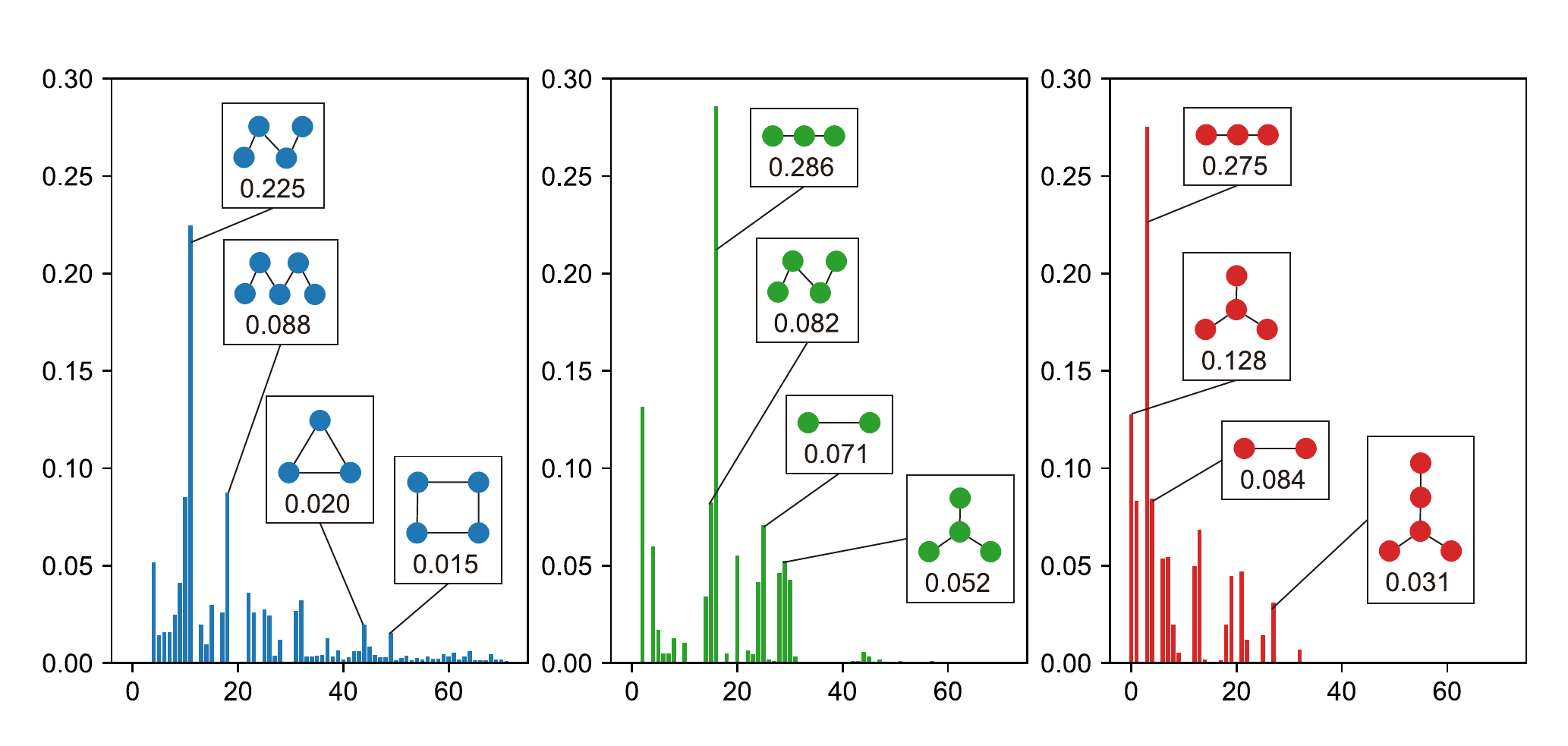}\label{fig:word-topic-anchor}}
         \subfigure[Walk-topic distribution by ordinary LDA]{ \includegraphics[width=0.46\linewidth]{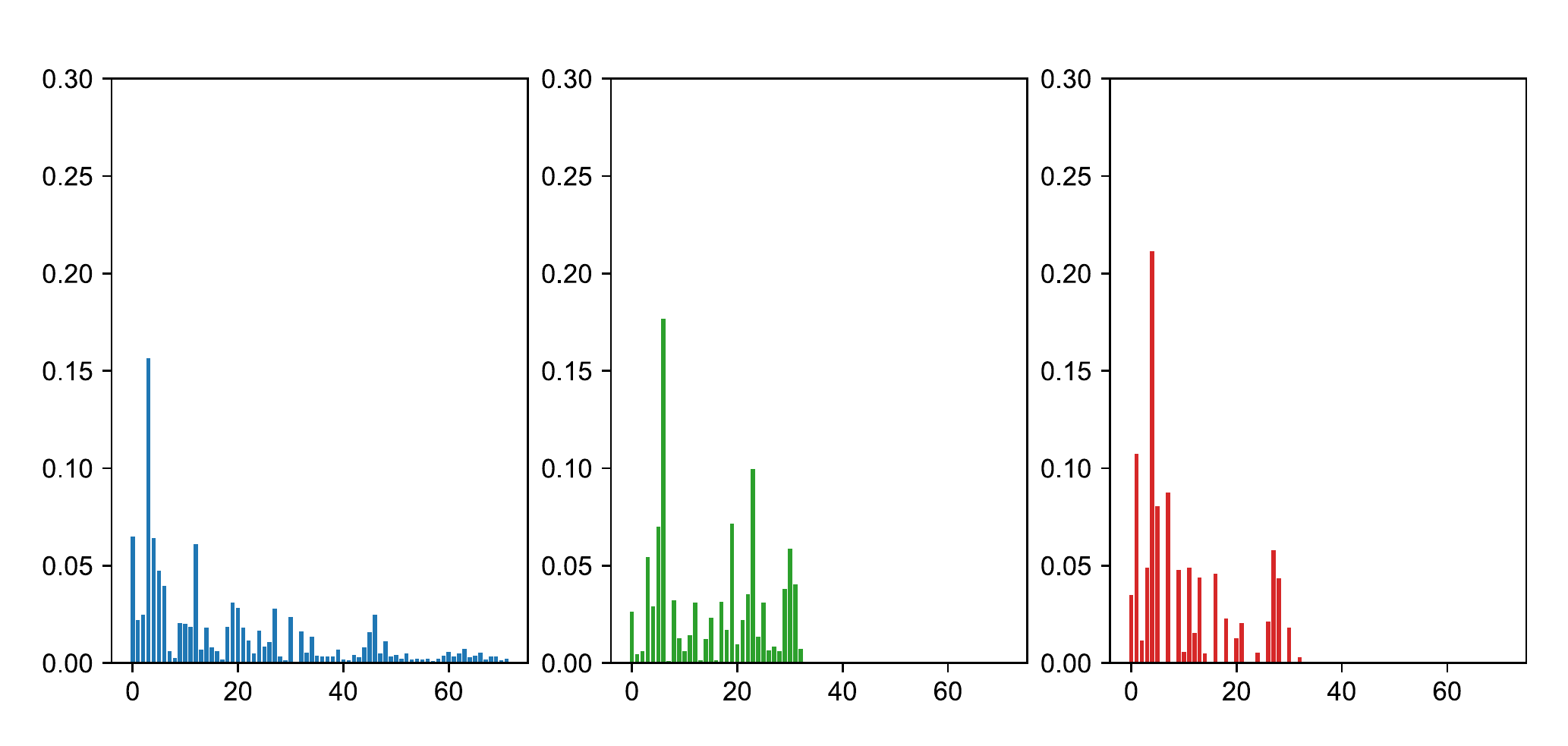}\label{fig:word-topic-ordinary}}
\caption{Visualization of walk-topic distributions by Graph Anchor LDA (left) and ordinary LDA (right). Graph Anchor LDA generates sharper walk-topic distributions, and amplifies indicative structural patterns within each structural topic. }
\label{fig:show_lda_distribution}
\end{figure*}

\section{Experiments}
In this section, we introduce our empirical evaluations of the model GraphSTONE. We first introduce experimental settings, before results on various tasks. Specifically, our evaluation consists of: 
\begin{itemize}
    \item Quantitative evaluation, including link reconstruction and node classification. 
    \item Qualitative evaluation, including visualization tasks that illustrate the results of Graph Anchor LDA and GraphSTONE in a straightforward manner. 
    \item Self evaluation, including analysis on model parameters and model components. 
\end{itemize}
\subsection{Experimental Setup}
We first introduce the datasets, comparison methods as well as settings for experimental evaluation. 

\noindent \textbf{Datasets} We utilize citation networks (Cora, Pubmed \cite{kipf2016semi}), social network (Aminer \cite{zhang2019dane}) and protein interaction network (PPI) to evaluate our model. Dataset statistics are listed in Table \ref{tab:datasets}. 

\begin{table}[htbp]
	\centering
	\begin{tabular}{ccccc}
	\toprule
	Datasets &
	{Type} &
	{$|V|$} &
	{$|E|$} &
	{\# Classes}\\
	\hline
	Cora & Citation & 2,708 & 5,429 & 7  \\
	AMiner & Social & 3,121 & 7,219 & 4 \\
	Pubmed & Citation & 19,717 & 44,338 & 3   \\
	PPI & Protein & 14,755 & 228,431 & 121   \\
	\bottomrule
    \end{tabular}
    \caption{Dataset statistics and properties.}
	\label{tab:datasets}
\end{table}

\noindent \textbf{Baselines} We take the following novel approaches in network representation learning as baselines.

\begin{itemize}
    \item \textbf{Structure models}, focusing on structural properties of nodes. Here we choose a popular model Struc2Vec \cite{ribeiro2017struc2vec}.
    \item \textbf{GNNs}, including GraphSAGE, GCN \cite{kipf2016semi} and GAT \cite{velivckovic2017graph}. We train these models using the unsupervised loss of Eq. \ref{eq:obj}. 
    \item \textbf{GraphSTONE (nf)}. We take the outputs of Graph Anchor LDA directly as inputs of GCN to verify how the extracted structural topics on networks contribute to better GCN modeling. We denote this variant as GraphSTONE (nf). Note that GraphSTONE (nf) does not take raw node features as inputs. 
\end{itemize}

\noindent \textbf{Settings} We take 64-dimensional embeddings for all methods, and adopt Adam optimizer with a learning rate of 0.005. For GNNs, we take 2-layer networks with a hidden layer sized 100. For skip-gram optimization (Eq. \ref{eq:obj}), we take $N = 100, l = 10$, window size as 5 and the number of negative sampling $q = 8$. For models involving neighborhood sampling, we take the number for sampling as 20. We leave the parameters of other baselines as default mentioned in corresponding papers. 
In addition, we take $K=5$ for GraphSTONE.

We also introduce two settings for node classification tasks. 
\begin{itemize}
\item \textbf{Transductive.} We allow all models access to the whole graph, i.e. all edges and node features. We apply this setting for Cora, AMiner, Pubmed and PPI. 

\item \textbf{Inductive.} The test nodes are unobserved during training. We apply this setting on PPI, where we train all GNNs on 20 graphs and directly predict on two fixed test graphs, as in \cite{hamilton2017inductive}. Note that only GNNs are capable of inductive classification. 
\end{itemize}

\begin{table*}
\centering
\begin{adjustbox}{max width=0.95\linewidth}
\begin{tabular}{llccccc c c c c}
\toprule
\multirow{2}{*}{Input}&
\multirow{2}{*}{Model}&
\multicolumn{2}{c}{Cora} &
\multicolumn{2}{c}{AMiner} &
\multicolumn{2}{c}{Pubmed}
\\
\cmidrule(lr){3-4} 
\cmidrule(lr){5-6} 
\cmidrule(lr){7-8} 
& & 
AUC & Recall@0.5 & AUC &  Recall@0.5  & AUC & 
Recall@0.5 \\

\midrule
\multirow{4}{*}{No features}
& Struc2Vec & 54.29 & 54.38 & 47.55 & 47.63 & 53.14 & 53.14 \\
& GraLSP & 66.28 & 66.38 & 65.40 & 65.50 & 57.62 & 57.63 \\
& GCN & 74.60 & 74.71 & 71.98 & 72.07 & 59.20 & 59.22\\
& GraphSTONE (nf) & \textbf{92.44} & \textbf{92.56} & \textbf{89.87} & \textbf{89.91} & \textbf{87.47} & \textbf{87.48} \\

\midrule
\multirow{5}{*}{Features}
& GCN & 94.14 & 94.26 & 94.47 & 94.55 & 92.23 & 92.25 \\
& GAT & 94.66 & 94.78 & 95.24 & 95.34 & 92.36 & 92.38 \\
& GraLSP & 94.39 & 94.51 & 94.85 & 94.89 & 90.83 & 90.84 \\
& GraphSAGE & 95.30 & 95.42 & 94.92 & 95.02 & 91.52 & 91.54 \\
& GraphSTONE & \textbf{96.37} & \textbf{96.70} & \textbf{95.94} & \textbf{96.06} & \textbf{94.25} & \textbf{94.27} \\
\bottomrule
\end{tabular}
\end{adjustbox}
\caption{Results of link reconstruction on different datasets.}
 \label{tab:link_pred}
\end{table*}

\begin{table*}[ht]
\centering
\begin{adjustbox}{max width=0.95\linewidth}
\begin{tabular}{llcccccccccccc c c c c c c c c c c c c c c c c c c c}
\toprule
\multirow{4}{*}{Input}&
\multirow{4}{*}{Model} &
\multicolumn{4}{c}{Cora} &
\multicolumn{4}{c}{AMiner} &
\multicolumn{4}{c}{Pubmed} & 
\multicolumn{4}{c}{PPI}&
\\
\cmidrule(lr){3-6} 
\cmidrule(lr){7-10} 
\cmidrule(lr){11-14}
\cmidrule(lr){15-18}
& &
\multicolumn{2}{c}{Macro-f1}& 
\multicolumn{2}{c}{Micro-f1}& 
\multicolumn{2}{c}{Macro-f1}& 
\multicolumn{2}{c}{Micro-f1}& 
\multicolumn{2}{c}{Macro-f1}& 
\multicolumn{2}{c}{Micro-f1}& 
\multicolumn{2}{c}{Macro-f1}& 
\multicolumn{2}{c}{Micro-f1}& 
\\
\cmidrule(lr){3-4} 
\cmidrule(lr){5-6}
\cmidrule(lr){7-8}
\cmidrule(lr){9-10}
\cmidrule(lr){11-12}
\cmidrule(lr){13-14}
\cmidrule(lr){15-16}
\cmidrule(lr){17-18}
& &
30\% & 70\% & 30\% & 70\% & 30\% & 70\% & 30\% & 70\% & 30\% & 70\% & 30\% & 70\% & 30\% & 70\% & 30\% & 70\%
\\
\midrule
\multirow{4}{*}{No features}
&Struc2Vec & 17.55 & 18.92 & 29.07 & 31.34 & 23.17 & 21.80 & 36.11 & 38.44 & 31.29 & 31.31 & 41.50 & 41.49 & 12.89 & 13.53 & 40.49 & 40.74\\
&GraLSP & 58.86 & 61.62 & 60.88 & 62.45 & 43.19 & 43.03 & 45.85 & 45.92 & 38.89 & 38.84 & 45.88 & 46.01 & 10.19 & 10.72 & 37.65 & 37.88 \\
&GCN & 11.65 & 11.94 & 32.30 & 32.83 & 14.86 & 16.81 & 41.24 & 42.51 & 35.07 & 36.51 & 46.56 & 47.83  & 8.75 & 9.08 & 36.70 & 37.46  \\
&GraphSTONE (nf) & \textbf{70.25} & \textbf{71.33} & \textbf{71.73} & \textbf{72.42} & \textbf{57.11}& \textbf{56.70} & \textbf{58.21}& \textbf{58.91} & \textbf{56.87}& \textbf{58.88} & \textbf{60.47}& \textbf{60.69} & 10.28 & 11.20 & 38.93 & 38.96  \\
\midrule
\multirow{5}{*}{Features}
&GCN & 79.84 & 81.09 & 80.97 & 81.94 & 65.02 & 67.33 & 64.89 & 66.72 & 76.93 & 77.21 & 76.42 & 77.49 & 12.57 & 12.62 & 40.40 & 40.44  \\
&GAT & 79.33 & 82.08 & 80.41 & 83.43  & 68.76& 69.10  & 67.92& 68.16 & 76.94& 76.92  & 77.64& 77.82 &  11.91 & 11.97 & 39.92 & 40.10 \\
&GraLSP & 82.43 & 83.27 & 83.67 & 84.31 & 68.82 & 70.15 & 69.12 & 69.73 & \textbf{81.21} & \textbf{81.38}  & \textbf{81.43} & \textbf{81.52} & 11.34 & 11.89 & 39.55 & 39.80  \\
&GraphSAGE & 80.52 & 81.90 & 82.13 & 83.17  & 67.40 & 68.32 & 66.59 & 67.54 & 76.61 & 77.24  & 77.36 & 77.84 & 11.81 & 12.41 & 39.80 &40.08  \\
&GraphSTONE & \textbf{82.78}& \textbf{83.54} & \textbf{83.88}& \textbf{84.73} & \textbf{69.37} & \textbf{71.16} & \textbf{69.51} & \textbf{69.93} & 78.61 & 78.87 & 79.53 & 81.03 & \textbf{15.55} & \textbf{15.91} & \textbf{43.60} & \textbf{43.64} 
\\
\bottomrule
\end{tabular}
\end{adjustbox}
\caption{Macro-f1 and Micro-f1 scores of transductive node classification.}
  \label{tab:classification}
\end{table*}

\subsection{Proof-of-concept Visualization}
As we propose a new problem -- topic modeling on graphs, we first show a visualization result to intuitively explain its results. We carry out visualization on a synthetic dataset $G(n)$ as a simple proof-of-concept. We design $G(n)$ using three types of structures: one dense cluster, one T-shaped and one star, and then connect $n$ such structures interleavingly on a ring. We show an illustration of $G(n)$ in Fig. \ref{fig:syn_original}. For a clear illustration, we replace each structure with a colored dot. Apparently, the nodes in $G(n)$ possess three distinct structural properties (with the ring excluded), which can be regarded as three structural topics. We then obtain representation vectors from GraphSTONE, GraLSP and MNMF, and plot them on a 2d plane. We also obtain a node-topic distribution with $K=3$ using Graph Anchor LDA and also plot them on the plane. 

As shown in Fig. \ref{fig:lda_anchor} and \ref{fig:lda_GraphSTONE}, both Graph Anchor LDA and GraphSTONE cluster the three types of nodes clearly. The results are even more astonishing as GraLSP (Fig. \ref{fig:lda_GraLSP}) fails to cluster nodes in a satisfactory manner, which shows that probabilistic topic modeling better captures indicative structural patterns and marks the difference between neighborhoods of nodes. Also, MNMF, a community-aware embedding algorithm, largely ignores the structural similarity between nodes and fails to separate nodes clearly. 

Moreover, we visualize the walk-topic distributions generated by Graph Anchor LDA in Fig. \ref{fig:word-topic-anchor}, and compare them with those by ordinary LDA, where x-axis denotes indices of anonymous walks, and y-axis denotes the corresponding probability. It can be shown that the anchors selected by our Graph Anchor LDA are not only indicative of ``topics'', but are also in accordance with the actual graph structures. Moreover, the walk-topic distributions generated by Graph Anchor LDA are indeed sharper than those by ordinary LDA, underscoring the need for selecting anchors. We hence conclude that Graph Anchor LDA is highly selective and interpretable in summarizing graph structures.
    
\subsection{Quantitative Evaluations}
We conduct experiments on link reconstruction and node classification to evaluate GraphSTONE quantitatively.

\subsubsection{Link Reconstruction}
 We conduct link reconstruction using Cora, AMiner, and Pubmed, but not PPI as it contains disjoint graphs. We sample a proportion of edges from the initial network, which are used as positive samples, along with an identical number of random negative edges. It should be noted that the positive edges \textit{are not} removed from training. We take the inner product of embedding vectors as the score for each sample, which is used for ranking. Samples with scores among top 50\% are predicted as ``positive'' edges. We report AUC and Recall as metrics. 
 
 The results are shown in Table \ref{tab:link_pred}. It can be showed that by combining structural features with original node features, our model is able to generate representations that more accurately summarizes the original network. It is even more remarkable that based on structural features alone, GraphSTONE(nf) is able to significantly outperform the counterparts without the help of raw node features. Although Struc2Vec considers global network structures, its poor performance in link reconstruction reveals its shortage in capturing the simplest structure -- edges. By comparison, GraphSTONE and GraphSTONE (nf) achieve a more elaborate balance among simple structures, like edges, and complex structures. 

\subsubsection{Node Classification}
We then carry out node classification according to the setting of each dataset. For the transductive setting, different fractions of nodes are sampled randomly for testing, leaving the rest for training. For the inductive setting, we take embedding vectors of nodes from the training graphs for training, and those from the test graphs for testing. We use Logistic Regression as the classifier and take the macro and micro F1 scores for evaluation. The results are averaged over 10 independent runs.

\noindent \textbf{Transductive classification} Results of transductive node classification are shown in Table \ref{tab:classification}. It is generally shown that compared to competitive baselines, GraphSTONE is able to achieve satisfactory results. In addition, GraphSTONE (nf) significantly outperforms GCN without feature input, hence verifying that the structural topic information extracted by topic modeling does contribute significantly to more accurate representations of nodes. 

\noindent \textbf{Inductive classification} Results of inductive classification are shown in Table \ref{tab:inductive}. Note that algorithms that do not refer to node features are not inductive and thus not listed. It is shown that GraphSTONE also outperforms competitive GNNs, showing that the learned structural topic features generalize well across networks of the same type. 

\begin{table}[t]
	\centering
	\begin{tabular}{ccccc}
	\toprule
	Model &
	Macro-f1 &
	Micro-f1 \\
	\hline
	GCN  & 12.15 & 40.85 \\
	GAT & 12.31 & 39.76 \\
	GraLSP  & 12.59 & 40.81\\
	GraphSAGE & 11.92  & 40.05  \\
	GraphSTONE & \textbf{18.14}  & \textbf{46.02} \\
	\bottomrule
    \end{tabular}
    \caption{Inductive node classification results on PPI.}
	\label{tab:inductive}
\end{table}

\subsection{Model Analysis}
We carry out model analysis, including ablation studies, parameter analysis and efficiency. We use node classification to reflect the performance of the model.

\subsubsection{Ablation studies}
We carry out thorough ablation studies on the two parts of GraphSTONE, topic modeling and GCN and show the results in Table \ref{tab:parameter}.

First, we delve deeper into the design of our Graph Anchor LDA by comparing it with several variants. On one hand, we vary the analogy of ``words'' by substituting anonymous walks with three other units of structures: individual nodes, distribution of node degrees in $D_i$ and random walks without anonymity. We denote the three variants as \textbf{GraphSTONE (node)}, \textbf{GraphSTONE (degree)} and \textbf{GraphSTONE (rw)}. On the other hand, we verify the contribution of ``anchors'' by comparing with GraphSTONE without anchors, referred to as \textbf{GraphSTONE (no anchors)}. It can be shown that all three variants, GraphSTONE (node), (degree) and (rw) all fail to outperform GraphSTONE, probably because individual nodes and degrees only depict one-hop structures, and raw random walks are too sparse to observe co-appearance. By comparison, GraphSTONE (no anchors) performs better than the previous three due to its use of anonymous walks, but worse than GraphSTONE, which endorses the contribution of anchors in topic modeling and corresponds with Fig. \ref{fig:word-topic-anchor}. 

\begin{table}[ht]
	\centering
	\begin{tabular}{ccc
	}
	\toprule
	Model & Macro-f1 & Micro-f1\\
	\hline
	GCN &  81.46 & 82.61  \\
	GraphSTONE (node) & 82.33 & 83.55 \\
	GraphSTONE (degree) & 81.29 & 82.36 \\
	GraphSTONE (rw) & 82.35 & 82.85 \\
	GraphSTONE (no anchors) & 83.12 & 83.39 \\
	GCN-concat & 81.43 & 82.62 \\
	\textbf{GraphSTONE} & \textbf{84.21} & \textbf{85.13} \\
	\bottomrule
    \end{tabular}
    \caption{Ablation studies of GraphSTONE.}
	\label{tab:parameter}
\end{table}

\begin{figure}[ht]
\centering
        \subfigure[Walk length $l$]{ \includegraphics[width=0.44\linewidth]{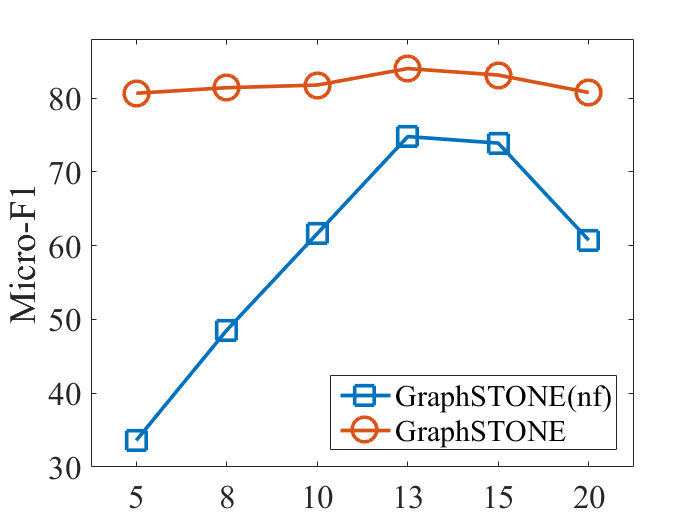}}
         \subfigure[Number of topics $K$]{ \includegraphics[width=0.44\linewidth]{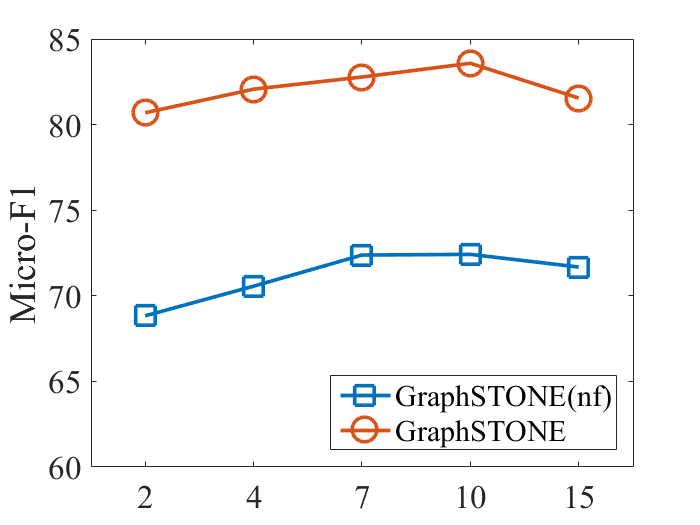}}
\caption{Parameter analysis of GraphSTONE.}
\label{fig:para_any}
\end{figure}

\begin{figure}[ht]
\centering
\includegraphics[width=0.85\linewidth]{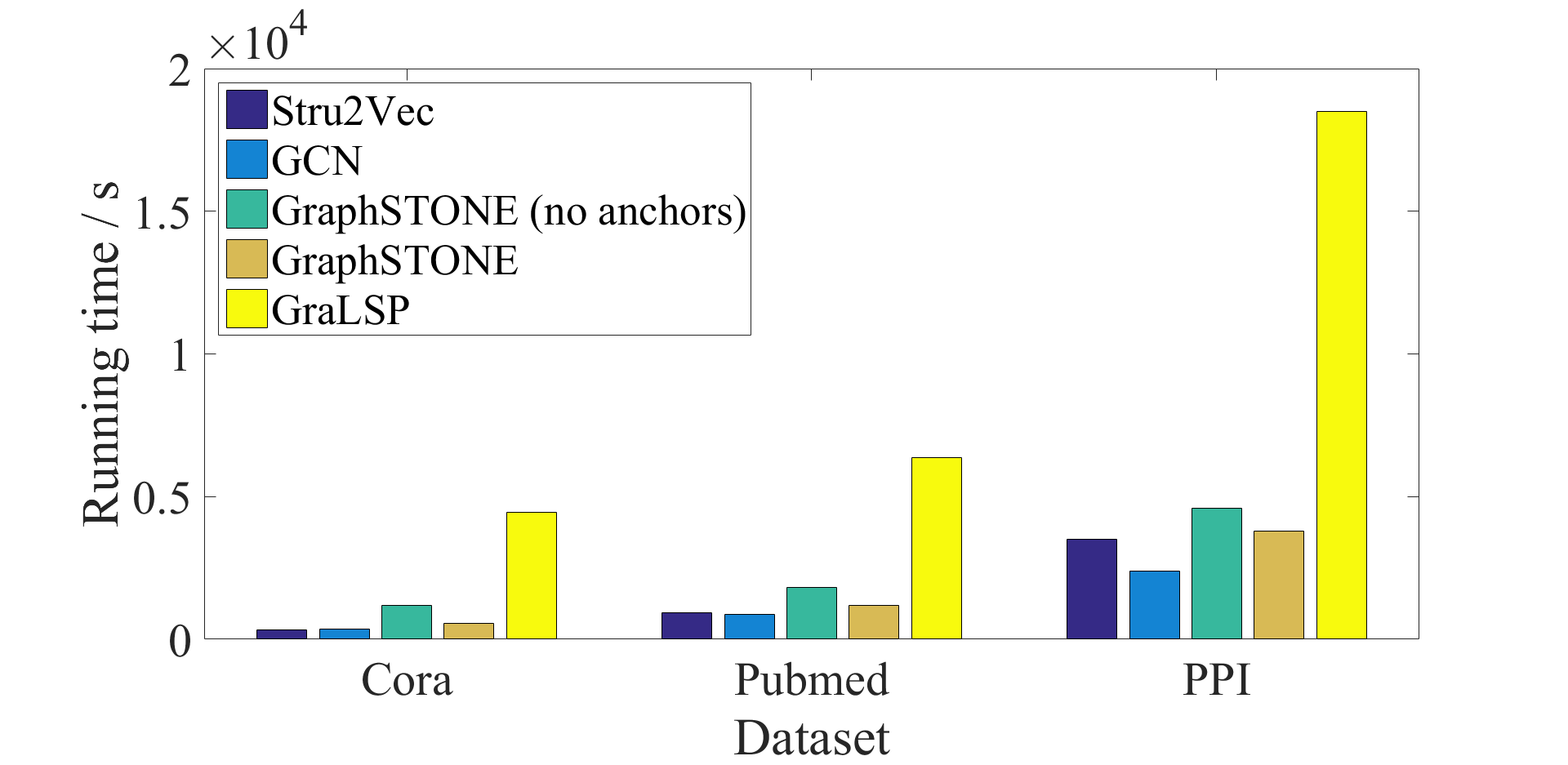}
\caption{Running time on different datasets.}
\label{fig:efficiency}
\end{figure}

\begin{figure*}[ht]
\centering
        \subfigure[GraphSAGE]{ \includegraphics[width=0.2\linewidth]{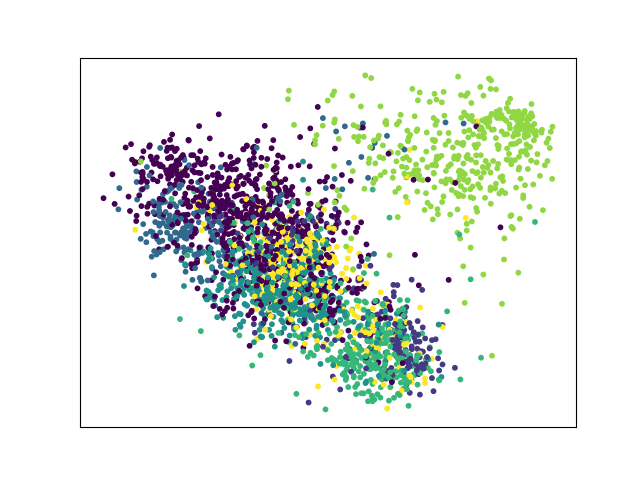}}
         \subfigure[Struc2Vec]{ \includegraphics[width=0.2\linewidth]{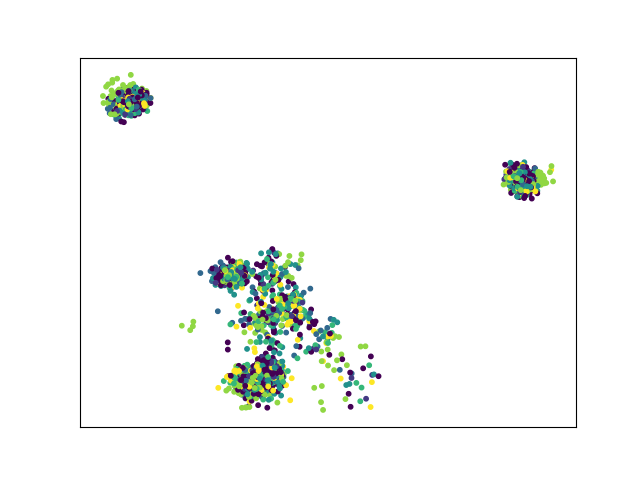}}
           \subfigure[GraLSP]{ \includegraphics[width=0.2\linewidth]{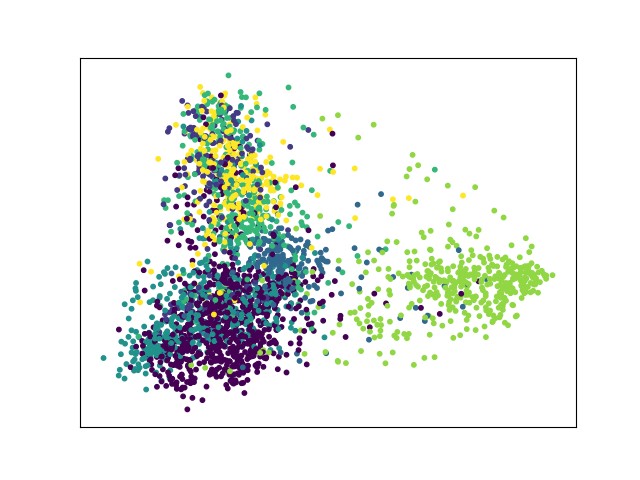}}
         \subfigure[GraphSTONE]{ \includegraphics[width=0.2\linewidth]{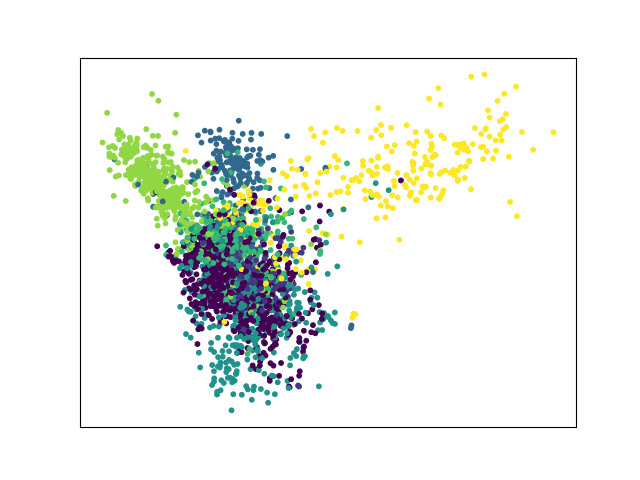}}
\caption{Visualization of representation vectors from various algorithms in 2D space.}
\label{fig:2d}
\end{figure*}

In addition, we validate the effectiveness of our structural multi-view GCN by comparing GraphSTONE with \textbf{GCN-concat}, a GCN with its input node features concatenated with structural topic features. It can be shown that with our designs of multi-view graph convolution, GraphSTONE enables node and structural features to complement each other, while simple aggregation fails to do so. 

\subsubsection{Parameter Analysis}
We analyze two parameters in our model, the length of walks $l$ and the number of topics $K$ and plot the performances in Fig. \ref{fig:para_any}. 
As shown, performances peak at certain choices for both parameters, while unduly set values, either too big or too small, will compromise the model performance. In addition, the sensitivity to parameter $l$ depends on whether to use node features. If node features are not used, GraphSTONE (nf) relies solely on structural topic features, thus being more sensitive to $l$. By comparison, GraphSTONE is less sensitive to $l$. 
\subsubsection{Efficiency}

We compared the efficiency of GraphSTONE with several baselines: Struc2Vec, GCN and GraLSP. We train all models and report the time needed for convergence on a single machine equipped with a GPU with 12GB RAM. 

Results are presented in Fig.\ref{fig:efficiency}. It can be shown that GraLSP suffers from poor efficiency due to its direct modeling of anonymous walks. By comparison, GraphSTONE takes significantly less time than GraLSP, showing the effectiveness of topic modeling. In addition, GraphSTONE becomes even more efficient when anchors are extracted and barely takes more time than GCN. 

\subsection{Visualization}
We carry out visualization on real-world datasets to qualitatively evaluate our model. We learn the representation vectors on Cora, which are then reduced to 2d vectors using PCA. We select representative models:
Struc2Vec (structure models), GraphSAGE (GNNs) and GraLSP (structure GNNs), and our model to compare.

The results are shown in Fig.\ref{fig:2d}, where different colors correspond to 7 labels in Cora. It is observed that GraphSTONE clusters nodes in a compact and separable manner, especially on certain colors (yellow, green and blue), compared with other methods.

\section{Conclusion}
In this paper, we present a GCN framework that captures local structural patterns for graph convolution, called GraphSTONE. To the best of our knowledge, it is the first attempt on topic modeling on graphs and GCNs. Specifically, we observe that the distributions, rather than individuals of local structural patterns are indicative towards node properties in networks, while current GCNs are scarcely capable of modeling. We then utilize topic modeling, specifically Graph Anchor LDA to capture the distributional differences over local structural patterns, and multi-view GCNs to incorporate such properties. We demonstrate that GraphSTONE is competitive, efficient and interpretable through multiple experiments. 

For future work, we seek to extend our work to see how GNNs are theoretically improved by incorporating various graph structures. 

\begin{acks}
We are grateful to Ziyao Li for his insightful advice towards this work. This work was supported by the National Natural Science Foundation of China (Grant No. 61876006 and No. 61572041).
\end{acks}

\bibliographystyle{ACM-Reference-Format}
\bibliography{kdd20}


\begin{thebibliography}{30}


\ifx \showCODEN    \undefined \def \showCODEN     #1{\unskip}     \fi
\ifx \showDOI      \undefined \def \showDOI       #1{#1}\fi
\ifx \showISBNx    \undefined \def \showISBNx     #1{\unskip}     \fi
\ifx \showISBNxiii \undefined \def \showISBNxiii  #1{\unskip}     \fi
\ifx \showISSN     \undefined \def \showISSN      #1{\unskip}     \fi
\ifx \showLCCN     \undefined \def \showLCCN      #1{\unskip}     \fi
\ifx \shownote     \undefined \def \shownote      #1{#1}          \fi
\ifx \showarticletitle \undefined \def \showarticletitle #1{#1}   \fi
\ifx \showURL      \undefined \def \showURL       {\relax}        \fi
\providecommand\bibfield[2]{#2}
\providecommand\bibinfo[2]{#2}
\providecommand\natexlab[1]{#1}
\providecommand\showeprint[2][]{arXiv:#2}

\bibitem[\protect\citeauthoryear{Arora, Ge, Halpern, Mimno, Moitra, Sontag, Wu,
  and Zhu}{Arora et~al\mbox{.}}{2013}]%
        {arora2013practical}
\bibfield{author}{\bibinfo{person}{Sanjeev Arora}, \bibinfo{person}{Rong Ge},
  \bibinfo{person}{Yonatan Halpern}, \bibinfo{person}{David Mimno},
  \bibinfo{person}{Ankur Moitra}, \bibinfo{person}{David Sontag},
  \bibinfo{person}{Yichen Wu}, {and} \bibinfo{person}{Michael Zhu}.}
  \bibinfo{year}{2013}\natexlab{}.
\newblock \showarticletitle{A practical algorithm for topic modeling with
  provable guarantees}. In \bibinfo{booktitle}{\emph{International Conference
  on Machine Learning}}. \bibinfo{pages}{280--288}.
\newblock


\bibitem[\protect\citeauthoryear{Arora, Ge, and Moitra}{Arora
  et~al\mbox{.}}{2012}]%
        {arora2012learning}
\bibfield{author}{\bibinfo{person}{Sanjeev Arora}, \bibinfo{person}{Rong Ge},
  {and} \bibinfo{person}{Ankur Moitra}.} \bibinfo{year}{2012}\natexlab{}.
\newblock \showarticletitle{Learning topic models--going beyond SVD}. In
  \bibinfo{booktitle}{\emph{2012 IEEE 53rd Annual Symposium on Foundations of
  Computer Science}}. IEEE, \bibinfo{pages}{1--10}.
\newblock


\bibitem[\protect\citeauthoryear{Blei, Ng, and Jordan}{Blei
  et~al\mbox{.}}{2003}]%
        {blei2003latent}
\bibfield{author}{\bibinfo{person}{David~M Blei}, \bibinfo{person}{Andrew~Y
  Ng}, {and} \bibinfo{person}{Michael~I Jordan}.}
  \bibinfo{year}{2003}\natexlab{}.
\newblock \showarticletitle{Latent dirichlet allocation}.
\newblock \bibinfo{journal}{\emph{Journal of machine Learning research}}
  \bibinfo{volume}{3}, \bibinfo{number}{Jan} (\bibinfo{year}{2003}),
  \bibinfo{pages}{993--1022}.
\newblock


\bibitem[\protect\citeauthoryear{Borgwardt and Kriegel}{Borgwardt and
  Kriegel}{2005}]%
        {borgwardt2005shortest}
\bibfield{author}{\bibinfo{person}{Karsten~M Borgwardt} {and}
  \bibinfo{person}{Hans-Peter Kriegel}.} \bibinfo{year}{2005}\natexlab{}.
\newblock \showarticletitle{Shortest-path kernels on graphs}. In
  \bibinfo{booktitle}{\emph{Fifth IEEE international conference on data
  mining}}. IEEE, \bibinfo{pages}{8--pp}.
\newblock


\bibitem[\protect\citeauthoryear{Donnat, Zitnik, Hallac, and Leskovec}{Donnat
  et~al\mbox{.}}{2018}]%
        {donnat2018learning}
\bibfield{author}{\bibinfo{person}{Claire Donnat}, \bibinfo{person}{Marinka
  Zitnik}, \bibinfo{person}{David Hallac}, {and} \bibinfo{person}{Jure
  Leskovec}.} \bibinfo{year}{2018}\natexlab{}.
\newblock \showarticletitle{Learning structural node embeddings via diffusion
  wavelets}. In \bibinfo{booktitle}{\emph{Proceedings of the 24th ACM SIGKDD
  International Conference on Knowledge Discovery \& Data Mining}}.
  \bibinfo{pages}{1320--1329}.
\newblock


\bibitem[\protect\citeauthoryear{Granovetter}{Granovetter}{1977}]%
        {granovetter1977strength}
\bibfield{author}{\bibinfo{person}{Mark~S Granovetter}.}
  \bibinfo{year}{1977}\natexlab{}.
\newblock \showarticletitle{The strength of weak ties}.
\newblock In \bibinfo{booktitle}{\emph{Social networks}}.
  \bibinfo{publisher}{Elsevier}, \bibinfo{pages}{347--367}.
\newblock


\bibitem[\protect\citeauthoryear{Hamilton, Ying, and Leskovec}{Hamilton
  et~al\mbox{.}}{2017}]%
        {hamilton2017inductive}
\bibfield{author}{\bibinfo{person}{Will Hamilton}, \bibinfo{person}{Zhitao
  Ying}, {and} \bibinfo{person}{Jure Leskovec}.}
  \bibinfo{year}{2017}\natexlab{}.
\newblock \showarticletitle{Inductive representation learning on large graphs}.
  In \bibinfo{booktitle}{\emph{Advances in Neural Information Processing
  Systems}}. \bibinfo{pages}{1024--1034}.
\newblock


\bibitem[\protect\citeauthoryear{Ivanov and Burnaev}{Ivanov and
  Burnaev}{2018}]%
        {ivanov2018anonymous}
\bibfield{author}{\bibinfo{person}{Sergey Ivanov} {and} \bibinfo{person}{Evgeny
  Burnaev}.} \bibinfo{year}{2018}\natexlab{}.
\newblock \showarticletitle{Anonymous Walk Embeddings}. In
  \bibinfo{booktitle}{\emph{International Conference on Machine Learning}}.
  \bibinfo{pages}{2191--2200}.
\newblock


\bibitem[\protect\citeauthoryear{Jin, You, Li, He, Cui, Fogelman-Souli{\'e},
  and Chakraborty}{Jin et~al\mbox{.}}{2019}]%
        {jin2019incorporating}
\bibfield{author}{\bibinfo{person}{Di Jin}, \bibinfo{person}{Xinxin You},
  \bibinfo{person}{Weihao Li}, \bibinfo{person}{Dongxiao He},
  \bibinfo{person}{Peng Cui}, \bibinfo{person}{Fran{\c{c}}oise
  Fogelman-Souli{\'e}}, {and} \bibinfo{person}{Tanmoy Chakraborty}.}
  \bibinfo{year}{2019}\natexlab{}.
\newblock \showarticletitle{Incorporating network embedding into markov random
  field for better community detection}. In
  \bibinfo{booktitle}{\emph{Proceedings of the AAAI Conference on Artificial
  Intelligence}}, Vol.~\bibinfo{volume}{33}. \bibinfo{pages}{160--167}.
\newblock


\bibitem[\protect\citeauthoryear{Jin, Song, and Shi}{Jin et~al\mbox{.}}{2020}]%
        {DBLP:conf/aaai/JinSS20}
\bibfield{author}{\bibinfo{person}{Yilun Jin}, \bibinfo{person}{Guojie Song},
  {and} \bibinfo{person}{Chuan Shi}.} \bibinfo{year}{2020}\natexlab{}.
\newblock \showarticletitle{GraLSP: Graph Neural Networks with Local Structural
  Patterns}. In \bibinfo{booktitle}{\emph{The Thirty-Fourth {AAAI} Conference
  on Artificial Intelligence, {AAAI} 2020, New York, NY, USA}}.
  \bibinfo{publisher}{{AAAI} Press}, \bibinfo{pages}{4361--4368}.
\newblock


\bibitem[\protect\citeauthoryear{Kawamae}{Kawamae}{2019}]%
        {kawamae2019topic}
\bibfield{author}{\bibinfo{person}{Noriaki Kawamae}.}
  \bibinfo{year}{2019}\natexlab{}.
\newblock \showarticletitle{Topic Structure-Aware Neural Language Model:
  Unified language model that maintains word and topic ordering by their
  embedded representations}. In \bibinfo{booktitle}{\emph{The World Wide Web
  Conference}}. ACM, \bibinfo{pages}{2900--2906}.
\newblock


\bibitem[\protect\citeauthoryear{Kipf and Welling}{Kipf and Welling}{2017}]%
        {kipf2016semi}
\bibfield{author}{\bibinfo{person}{Thomas Kipf} {and} \bibinfo{person}{Max
  Welling}.} \bibinfo{year}{2017}\natexlab{}.
\newblock \showarticletitle{Semi-Supervised Classification with Graph
  Convolutional Networks}. In \bibinfo{booktitle}{\emph{International
  Conference of Learning Representations}}.
\newblock


\bibitem[\protect\citeauthoryear{Koutra, Kang, Vreeken, and Faloutsos}{Koutra
  et~al\mbox{.}}{2014}]%
        {koutra2014vog}
\bibfield{author}{\bibinfo{person}{Danai Koutra}, \bibinfo{person}{U Kang},
  \bibinfo{person}{Jilles Vreeken}, {and} \bibinfo{person}{Christos
  Faloutsos}.} \bibinfo{year}{2014}\natexlab{}.
\newblock \showarticletitle{Vog: Summarizing and understanding large graphs}.
  In \bibinfo{booktitle}{\emph{Proceedings of the 2014 SIAM international
  conference on data mining}}. SIAM, \bibinfo{pages}{91--99}.
\newblock


\bibitem[\protect\citeauthoryear{Lee and Seung}{Lee and Seung}{1999}]%
        {1999learning}
\bibfield{author}{\bibinfo{person}{Daniel~D Lee} {and}
  \bibinfo{person}{H~Sebastian Seung}.} \bibinfo{year}{1999}\natexlab{}.
\newblock \showarticletitle{Learning the parts of objects by non-negative
  matrix factorization}.
\newblock \bibinfo{journal}{\emph{Nature}} \bibinfo{volume}{401},
  \bibinfo{number}{6755} (\bibinfo{year}{1999}), \bibinfo{pages}{788}.
\newblock


\bibitem[\protect\citeauthoryear{Lee, Rossi, Kong, Kim, Koh, and Rao}{Lee
  et~al\mbox{.}}{2019}]%
        {lee2019graph}
\bibfield{author}{\bibinfo{person}{John~Boaz Lee}, \bibinfo{person}{Ryan~A
  Rossi}, \bibinfo{person}{Xiangnan Kong}, \bibinfo{person}{Sungchul Kim},
  \bibinfo{person}{Eunyee Koh}, {and} \bibinfo{person}{Anup Rao}.}
  \bibinfo{year}{2019}\natexlab{}.
\newblock \showarticletitle{Graph Convolutional Networks with Motif-based
  Attention}. In \bibinfo{booktitle}{\emph{Proceedings of the 28th ACM
  International Conference on Information and Knowledge Management}}.
  \bibinfo{pages}{499--508}.
\newblock


\bibitem[\protect\citeauthoryear{Li, Zhang, and Song}{Li et~al\mbox{.}}{2019}]%
        {li2019gcn}
\bibfield{author}{\bibinfo{person}{Ziyao Li}, \bibinfo{person}{Liang Zhang},
  {and} \bibinfo{person}{Guojie Song}.} \bibinfo{year}{2019}\natexlab{}.
\newblock \showarticletitle{GCN-LASE: towards adequately incorporating link
  attributes in graph convolutional networks}. In
  \bibinfo{booktitle}{\emph{Proceedings of the 28th International Joint
  Conference on Artificial Intelligence}}. AAAI Press,
  \bibinfo{pages}{2959--2965}.
\newblock


\bibitem[\protect\citeauthoryear{Liu, Tang, He, Yao, and Zhou}{Liu
  et~al\mbox{.}}{2017}]%
        {liu2017predicting}
\bibfield{author}{\bibinfo{person}{Lin Liu}, \bibinfo{person}{Lin Tang},
  \bibinfo{person}{Libo He}, \bibinfo{person}{Shaowen Yao}, {and}
  \bibinfo{person}{Wei Zhou}.} \bibinfo{year}{2017}\natexlab{}.
\newblock \showarticletitle{Predicting protein function via multi-label
  supervised topic model on gene ontology}.
\newblock \bibinfo{journal}{\emph{Biotechnology \& Biotechnological Equipment}}
  \bibinfo{volume}{31}, \bibinfo{number}{3} (\bibinfo{year}{2017}),
  \bibinfo{pages}{630--638}.
\newblock


\bibitem[\protect\citeauthoryear{Liu, Liu, Chua, and Sun}{Liu
  et~al\mbox{.}}{2015}]%
        {liu2015topical}
\bibfield{author}{\bibinfo{person}{Yang Liu}, \bibinfo{person}{Zhiyuan Liu},
  \bibinfo{person}{Tat-Seng Chua}, {and} \bibinfo{person}{Maosong Sun}.}
  \bibinfo{year}{2015}\natexlab{}.
\newblock \showarticletitle{Topical word embeddings}. In
  \bibinfo{booktitle}{\emph{Twenty-Ninth AAAI Conference on Artificial
  Intelligence}}.
\newblock


\bibitem[\protect\citeauthoryear{Long, Wang, Du, Song, Jin, and Lin}{Long
  et~al\mbox{.}}{2019}]%
        {long2019hierarchical}
\bibfield{author}{\bibinfo{person}{Qingqing Long}, \bibinfo{person}{Yiming
  Wang}, \bibinfo{person}{Lun Du}, \bibinfo{person}{Guojie Song},
  \bibinfo{person}{Yilun Jin}, {and} \bibinfo{person}{Wei Lin}.}
  \bibinfo{year}{2019}\natexlab{}.
\newblock \showarticletitle{Hierarchical Community Structure Preserving Network
  Embedding: A Subspace Approach}. In \bibinfo{booktitle}{\emph{Proceedings of
  the 28th ACM International Conference on Information and Knowledge
  Management}}. \bibinfo{pages}{409--418}.
\newblock


\bibitem[\protect\citeauthoryear{Loukas}{Loukas}{2020}]%
        {Loukas2020What}
\bibfield{author}{\bibinfo{person}{Andreas Loukas}.}
  \bibinfo{year}{2020}\natexlab{}.
\newblock \showarticletitle{What graph neural networks cannot learn: depth vs
  width}. In \bibinfo{booktitle}{\emph{International Conference on Learning
  Representations}}.
\newblock
\urldef\tempurl%
\url{https://openreview.net/forum?id=B1l2bp4YwS}
\showURL{%
\tempurl}


\bibitem[\protect\citeauthoryear{Micali and Zhu}{Micali and Zhu}{2016}]%
        {micali2016reconstructing}
\bibfield{author}{\bibinfo{person}{Silvio Micali} {and}
  \bibinfo{person}{Zeyuan~Allen Zhu}.} \bibinfo{year}{2016}\natexlab{}.
\newblock \showarticletitle{Reconstructing markov processes from independent
  and anonymous experiments}.
\newblock \bibinfo{journal}{\emph{Discrete Applied Mathematics}}
  \bibinfo{volume}{200} (\bibinfo{year}{2016}), \bibinfo{pages}{108--122}.
\newblock


\bibitem[\protect\citeauthoryear{Milo, Shen-Orr, Itzkovitz, Kashtan,
  Chklovskii, and Alon}{Milo et~al\mbox{.}}{2002}]%
        {milo2002network}
\bibfield{author}{\bibinfo{person}{Ron Milo}, \bibinfo{person}{Shai Shen-Orr},
  \bibinfo{person}{Shalev Itzkovitz}, \bibinfo{person}{Nadav Kashtan},
  \bibinfo{person}{Dmitri Chklovskii}, {and} \bibinfo{person}{Uri Alon}.}
  \bibinfo{year}{2002}\natexlab{}.
\newblock \showarticletitle{Network motifs: simple building blocks of complex
  networks}.
\newblock \bibinfo{journal}{\emph{Science}} \bibinfo{volume}{298},
  \bibinfo{number}{5594} (\bibinfo{year}{2002}), \bibinfo{pages}{824--827}.
\newblock


\bibitem[\protect\citeauthoryear{Morris, Ritzert, Fey, Hamilton, Lenssen,
  Rattan, and Grohe}{Morris et~al\mbox{.}}{2019}]%
        {morris2019weisfeiler}
\bibfield{author}{\bibinfo{person}{Christopher Morris}, \bibinfo{person}{Martin
  Ritzert}, \bibinfo{person}{Matthias Fey}, \bibinfo{person}{William~L
  Hamilton}, \bibinfo{person}{Jan~Eric Lenssen}, \bibinfo{person}{Gaurav
  Rattan}, {and} \bibinfo{person}{Martin Grohe}.}
  \bibinfo{year}{2019}\natexlab{}.
\newblock \showarticletitle{Weisfeiler and leman go neural: Higher-order graph
  neural networks}. In \bibinfo{booktitle}{\emph{Proceedings of the AAAI
  Conference on Artificial Intelligence}}, Vol.~\bibinfo{volume}{33}.
  \bibinfo{pages}{4602--4609}.
\newblock


\bibitem[\protect\citeauthoryear{Oono and Suzuki}{Oono and Suzuki}{2020}]%
        {oono2020graph}
\bibfield{author}{\bibinfo{person}{Kenta Oono} {and} \bibinfo{person}{Taiji
  Suzuki}.} \bibinfo{year}{2020}\natexlab{}.
\newblock \showarticletitle{Graph Neural Networks Exponentially Lose Expressive
  Power for Node Classification}. In \bibinfo{booktitle}{\emph{International
  Conference on Learning Representations}}.
\newblock
\urldef\tempurl%
\url{https://openreview.net/forum?id=S1ldO2EFPr}
\showURL{%
\tempurl}


\bibitem[\protect\citeauthoryear{Ribeiro, Saverese, and Figueiredo}{Ribeiro
  et~al\mbox{.}}{2017}]%
        {ribeiro2017struc2vec}
\bibfield{author}{\bibinfo{person}{Leonardo~FR Ribeiro},
  \bibinfo{person}{Pedro~HP Saverese}, {and} \bibinfo{person}{Daniel~R
  Figueiredo}.} \bibinfo{year}{2017}\natexlab{}.
\newblock \showarticletitle{struc2vec: Learning node representations from
  structural identity}. In \bibinfo{booktitle}{\emph{Proceedings of the 23rd
  ACM SIGKDD International Conference on Knowledge Discovery and Data Mining}}.
  ACM, \bibinfo{pages}{385--394}.
\newblock


\bibitem[\protect\citeauthoryear{Shervashidze, Vishwanathan, Petri, Mehlhorn,
  and Borgwardt}{Shervashidze et~al\mbox{.}}{2009}]%
        {shervashidze2009efficient}
\bibfield{author}{\bibinfo{person}{Nino Shervashidze}, \bibinfo{person}{SVN
  Vishwanathan}, \bibinfo{person}{Tobias Petri}, \bibinfo{person}{Kurt
  Mehlhorn}, {and} \bibinfo{person}{Karsten Borgwardt}.}
  \bibinfo{year}{2009}\natexlab{}.
\newblock \showarticletitle{Efficient graphlet kernels for large graph
  comparison}. In \bibinfo{booktitle}{\emph{Artificial Intelligence and
  Statistics}}. \bibinfo{pages}{488--495}.
\newblock


\bibitem[\protect\citeauthoryear{Veli{\v{c}}kovi{\'c}, Cucurull, Casanova,
  Romero, Lio, and Bengio}{Veli{\v{c}}kovi{\'c} et~al\mbox{.}}{2017}]%
        {velivckovic2017graph}
\bibfield{author}{\bibinfo{person}{Petar Veli{\v{c}}kovi{\'c}},
  \bibinfo{person}{Guillem Cucurull}, \bibinfo{person}{Arantxa Casanova},
  \bibinfo{person}{Adriana Romero}, \bibinfo{person}{Pietro Lio}, {and}
  \bibinfo{person}{Yoshua Bengio}.} \bibinfo{year}{2017}\natexlab{}.
\newblock \showarticletitle{Graph attention networks}.
\newblock \bibinfo{journal}{\emph{arXiv preprint arXiv:1710.10903}}
  (\bibinfo{year}{2017}).
\newblock


\bibitem[\protect\citeauthoryear{Xu, Hu, Leskovec, and Jegelka}{Xu
  et~al\mbox{.}}{2018}]%
        {xu2018powerful}
\bibfield{author}{\bibinfo{person}{Keyulu Xu}, \bibinfo{person}{Weihua Hu},
  \bibinfo{person}{Jure Leskovec}, {and} \bibinfo{person}{Stefanie Jegelka}.}
  \bibinfo{year}{2018}\natexlab{}.
\newblock \showarticletitle{How powerful are graph neural networks?}
\newblock \bibinfo{journal}{\emph{arXiv preprint arXiv:1810.00826}}
  (\bibinfo{year}{2018}).
\newblock


\bibitem[\protect\citeauthoryear{Zhang, Song, Du, Yang, and Jin}{Zhang
  et~al\mbox{.}}{2019}]%
        {zhang2019dane}
\bibfield{author}{\bibinfo{person}{Yizhou Zhang}, \bibinfo{person}{Guojie
  Song}, \bibinfo{person}{Lun Du}, \bibinfo{person}{Shuwen Yang}, {and}
  \bibinfo{person}{Yilun Jin}.} \bibinfo{year}{2019}\natexlab{}.
\newblock \showarticletitle{DANE: Domain Adaptive Network Embedding}. In
  \bibinfo{booktitle}{\emph{Proceedings of the Twenty-Eighth International
  Joint Conference on Artificial Intelligence}}.
\newblock


\bibitem[\protect\citeauthoryear{Zhou, Yang, Ren, Wu, and Zhuang}{Zhou
  et~al\mbox{.}}{2018}]%
        {zhou2018dynamic}
\bibfield{author}{\bibinfo{person}{Lekui Zhou}, \bibinfo{person}{Yang Yang},
  \bibinfo{person}{Xiang Ren}, \bibinfo{person}{Fei Wu}, {and}
  \bibinfo{person}{Yueting Zhuang}.} \bibinfo{year}{2018}\natexlab{}.
\newblock \showarticletitle{Dynamic network embedding by modeling triadic
  closure process}. In \bibinfo{booktitle}{\emph{Thirty-Second AAAI Conference
  on Artificial Intelligence}}.
\newblock


\end{thebibliography}
\end{document}